\documentclass[11pt]{searcheyes}

\usepackage{enumitem}
\usepackage{tabularx}
\usepackage{float}

\newfontfamily\arxivkeepbytesansmedium[Path=./]{ByteSans-Medium.ttf}
\newfontfamily\arxivkeepbytesansbold[Path=./]{ByteSans-Bold.ttf}
\newsavebox{\arxivdependencybox}
\AtBeginDocument{%
  \sbox{\arxivdependencybox}{%
    \includegraphics{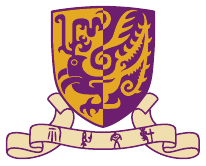}%
    \includegraphics{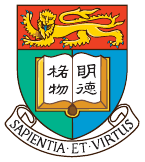}%
    \includegraphics{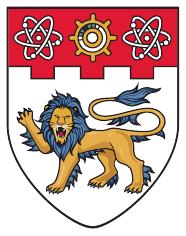}%
    \includegraphics{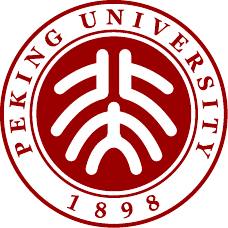}%
    \includegraphics{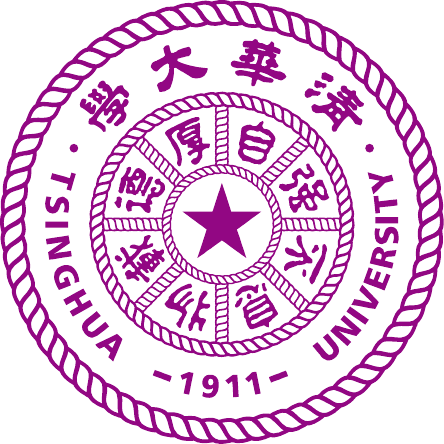}%
  }%
}

\title{\raisebox{-0.3em}{\includegraphics[height=1.5em]{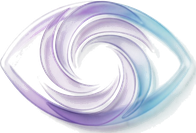}}\kern0.4em \gradtext{SearchEyes}: Towards Frontier Multimodal Deep Search Intelligence via Search World Simulation}

\author[1]{Zhengbo Jiao}
\author[2]{Yiming Cheng}
\author[1]{Yilei Jiang}
\author[1]{Kaituo Feng}
\author[3]{Rui Huang}
\author[4]{\\Tianyi Jiang}
\author[5]{Juanxi Tian}
\author[2]{Jiapeng Li}
\author[1]{Qunzhong Wang}
\author[6]{Tailai Chen}
\author[6]{\\Qianshan Wei}
\author[8]{Chuan Xiao}
\author[4]{Shanyu Rong}
\author[7]{Yangfu Li}
\author[2]{Yanhan Zhou}
\author[9]{\\Yunpu Ma}
\author[6]{Yifan Zhang}
\author[1,\dagger]{Xiangyu Yue}

\affiliation[1]{MMLab, CUHK}
\affiliation[2]{THU}
\affiliation[3]{HKU}
\affiliation[4]{PKU}
\affiliation[5]{NTU}
\affiliation[6]{CASIA}
\affiliation[7]{ECNU}
\affiliation[8]{HIT}
\affiliation[9]{LMU}

\abstract{%
Training multimodal search agents to perform multi-hop reasoning remains challenging due to a fundamental structural disconnect: existing pipelines construct training data, search environments, and reward signals independently, causing synthesized structural metadata to be discarded, environments to rely on irreproducible external engines, and RL rewards to remain sparse at the trajectory level. We present \textbf{SearchEyes}, which uses a typed knowledge graph as the backbone of a \emph{simulated search world} that unifies all three components. We propose \textbf{Perception-Knowledge Chains (PKC)} to sample constrained multi-hop paths over the visual-knowledge intersection of Wikidata5M, retaining hop-level entity metadata that simultaneously defines a self-contained search world and step-level reward anchors. We further propose \textbf{Hop-Anchored Policy Optimization (HaPO)}, which reuses these anchors for step-level credit assignment without a separately trained process reward model. Experiments on six multimodal knowledge-intensive benchmarks show that SearchEyes achieves state-of-the-art performance among open-source multimodal search agents, with SearchEyes-27B improving over the strongest open-source baseline by 6.2 points on average.%
}

\date{June 2026}
\correspondence{{\email{Frostlin2005@gmail.com};}\email{xyyue@ie.cuhk.edu.hk}}
\checkdata[Project Page]{\url{https://github.com/Frostlinx/SearchEyes}}

\begin{document}
\maketitle

% ── Sections ─────────────────────────────────────
% ══════════════════════════════════════════════════
% 1. Introduction
% ══════════════════════════════════════════════════
\section{Introduction}

\begin{figure*}[tb!]
  \centering
  \includegraphics[width=\textwidth]{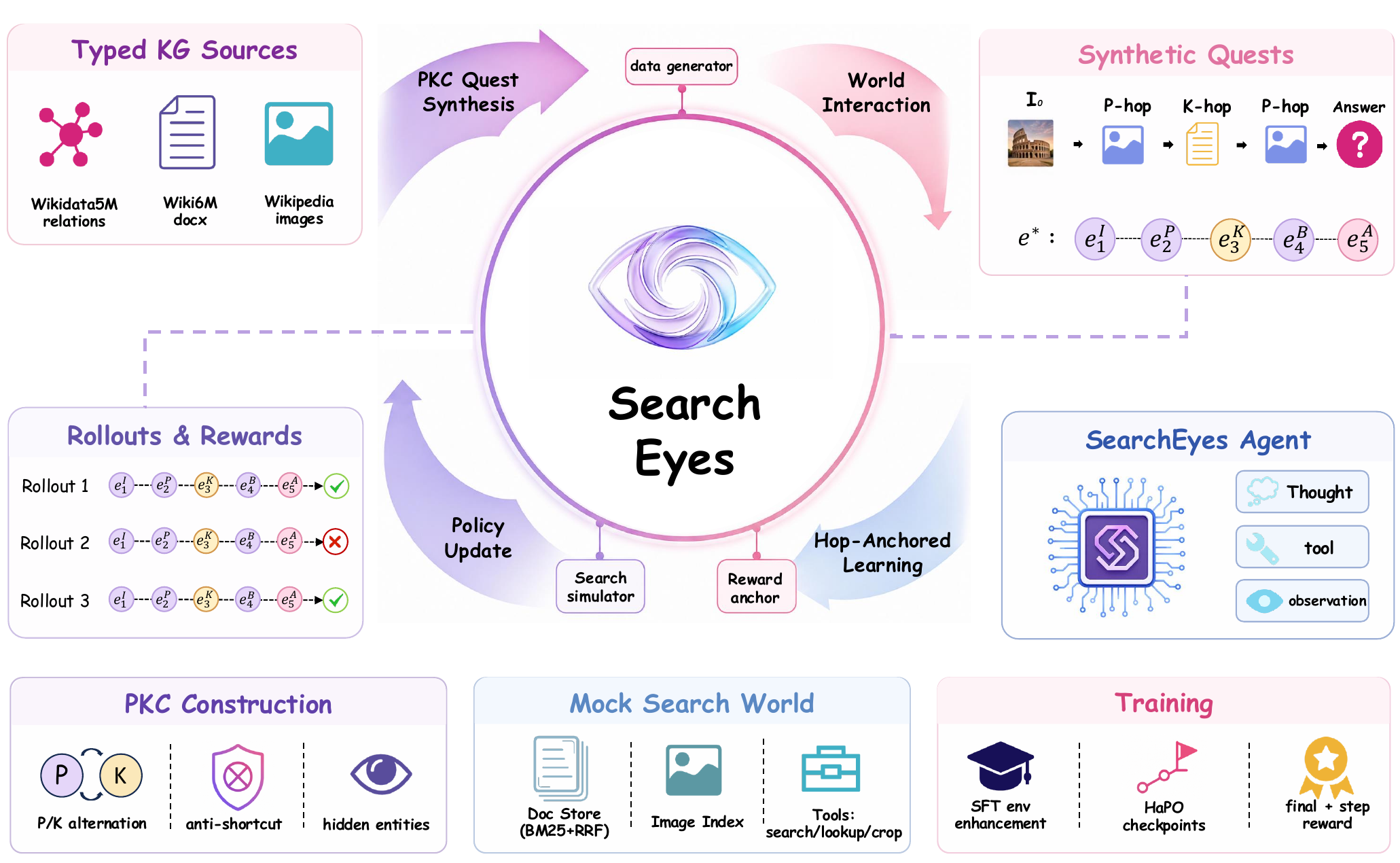}
  \caption{\textbf{Overview of SearchEyes.} A typed knowledge graph serves as the unified backbone for three tightly-coupled stages: PKC question synthesis (left), a self-contained search world for agent interaction (center), and hop-anchored reward signals for policy optimization (right). By retaining structural metadata across all stages, SearchEyes eliminates the data--environment--reward disconnect that limits existing pipelines.}
  \label{fig:overview}
\end{figure*}

% ¶1 Background
Reinforcement learning has become the dominant paradigm for training autonomous LLM agents~\cite{kimik25, glm5, qwen35, deepseekv4}, enabling capabilities that span long-chain reasoning, tool use, and multi-step decision making.
Among these, multi-step information retrieval and synthesis has emerged as a particularly impactful application, where agents iteratively search, read, and reason over external knowledge to solve complex questions.
On the text side, systems such as Search-R1~\cite{jin2025searchr1}, ASearcher~\cite{gao2025asearcher}, and REDSearcher~\cite{chu2026redsearcher} have demonstrated that RL-trained search agents can effectively perform long-horizon reasoning over retrieved evidence.
In the multimodal domain, Vision-DeepResearch~\cite{huang2026visiondeepresearch}, OpenSearch-VL~\cite{chen2026opensearchvl}, MMSearch-R1~\cite{wu2026mmsearchr1}, and VSearcher~\cite{zhang2026vsearcher} have begun equipping MLLMs with visual and textual search tools, coupled with dedicated training environments and agentic RL algorithms.
While these efforts have yielded impressive results, training a strong long-horizon multimodal search agent remains an open challenge, as current pipelines exhibit fundamental bottlenecks in how training data, search environments, and reward signals are constructed and connected.

% ¶2 Motivation
Despite the promising results above, existing pipelines suffer from a fundamental structural disconnect: the training data, search environment, and reward signal are constructed independently, with no shared structure binding them.
On the data side, prior methods leverage graph structures to synthesize multi-hop questions~\cite{chen2026opensearchvl, huang2026visiondeepresearch, deepdive2025}, yet discard all intermediate metadata (path structure, entity IDs, relation types) after synthesis, delivering only question-answer pairs to the training stage.
This decoupling propagates to the environment: without shared structural information, existing systems resort to external search engines as training environments~\cite{jin2025searchr1, chu2026redsearcher, li2025webthinker}, introducing irreproducibility, nondeterminism, and high API costs.
The consequences are most acute at the RL level, where reward signals can only be computed at the trajectory level based on final answers~\cite{grpo, dapo}.
For long-horizon multi-hop reasoning, the probability of completing an entire chain correctly diminishes sharply with the number of hops, rendering trajectory-level rewards prohibitively sparse for effective policy optimization~\cite{gigpo, arpo, prime}.

% ¶3 Solution
To address this structural disconnect, a key observation is that a typed knowledge graph can simultaneously serve as the structural backbone for all three components, effectively functioning as a \emph{simulated search world}.
We present \textbf{SearchEyes}, which uses the typed triples of Wikidata5M to construct such a search world that unifies data synthesis, environment simulation, and RL training.
On the data side, we propose \textbf{Perception-Knowledge Chains (PKC)}, which sample constrained multi-hop paths over the intersection of visual and knowledge entities, followed by multi-level anti-shortcut filtering to produce high-quality training data.
Since all entities reside in the same knowledge base, this base naturally constitutes a self-contained search world without external APIs.
On the training side, we propose \textbf{Hop-Anchored Policy Optimization (HaPO)}, which reuses the gold entity IDs retained by PKC as step-level anchors for credit assignment, blending hop-anchored and trajectory-level advantages without training a separate process reward model.

% ¶4 Experimental highlights
Figure~\ref{fig:overview} illustrates the overall framework.
Experiments on six multimodal knowledge-intensive benchmarks demonstrate that SearchEyes achieves state-of-the-art performance among open-source multimodal search agents; our SearchEyes-27B model improves over the strongest open-source baseline by 6.2 points on average.
We will release all data, code, and model weights to facilitate future research.

% ¶5 Contributions
Our contributions are summarized as follows:
\begin{enumerate}[leftmargin=1.5em]
  \item \textbf{Unified Search World Simulation.} We introduce SearchEyes, which uses the typed triples of Wikidata5M to construct a simulated search world that unifies data synthesis, environment simulation, and RL training, achieving data-environment-algorithm co-design.

  \item \textbf{Perception-Knowledge Chain Synthesis.} We propose PKC, a KG-driven pipeline that samples constrained multi-hop paths over the visual-knowledge intersection graph, with multi-level anti-shortcut filtering to produce high-quality training data. KG metadata is retained throughout synthesis to support downstream step-level training.

  \item \textbf{Agentic Post-Training Pipeline.} We design a complete post-training pipeline comprising SFT with hinted environments and observation denoising, followed by HaPO (Hop-Anchored Policy Optimization), which reuses PKC's hop-level gold entity IDs for step-level credit assignment without a separately trained process reward model.

  \item \textbf{Empirical Results.} Experiments on six multimodal knowledge-intensive benchmarks show that SearchEyes achieves state-of-the-art performance among open-source multimodal search agents; SearchEyes-27B improves over the strongest open-source baseline by 6.2 points on average. We additionally introduce VisSearch Bench for multi-hop visual search evaluation (Appendix~\ref{app:vissearch}).
\end{enumerate}

\begin{figure*}[tb!]
  \centering
  \includegraphics[width=\textwidth]{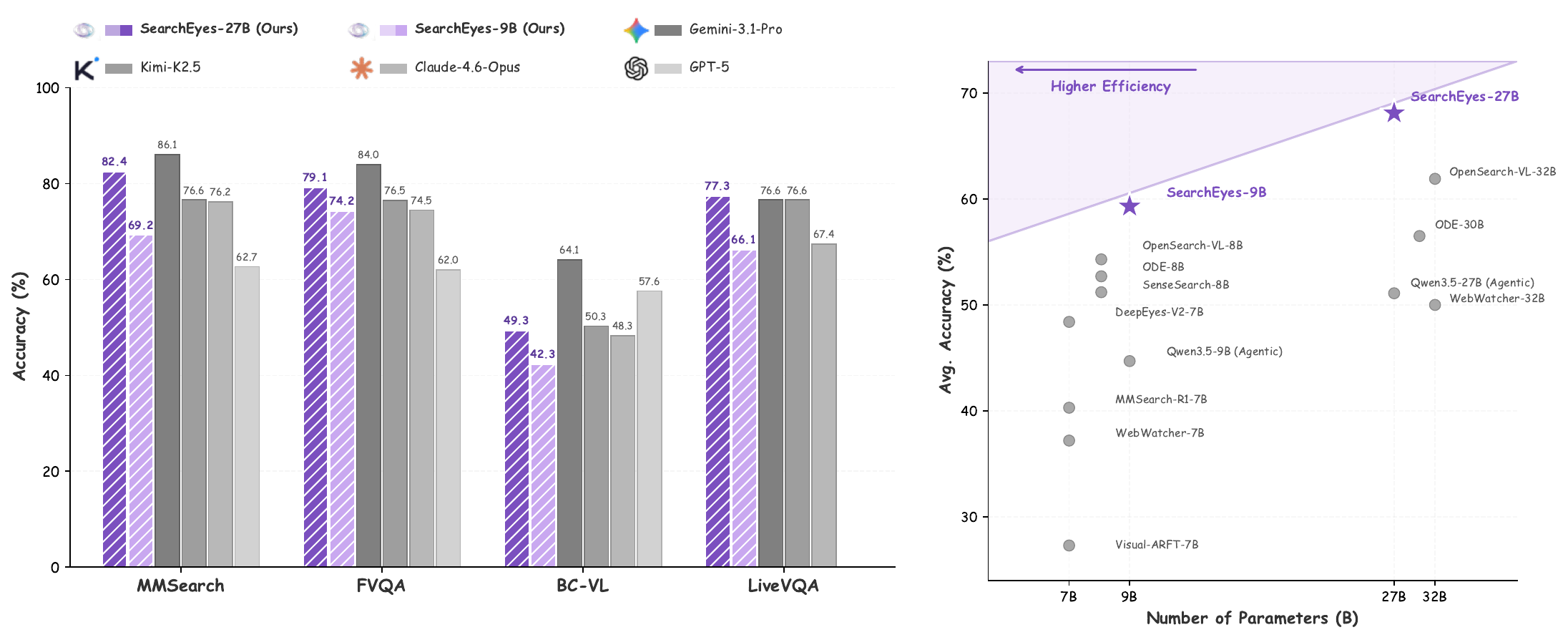}
  \caption{\textbf{Comparison of model scale versus average accuracy on multimodal knowledge-intensive benchmarks.} \emph{Left:} SearchEyes-27B (purple, hatched) achieves 82.4 on MMSearch and 79.1 on FVQA, competitive with Gemini-3.1-Pro while using only a 27B open-source backbone. \emph{Right:} Parameter efficiency of open-source multimodal search agents. SearchEyes-9B matches 30B-scale baselines (e.g., OpenSearch-VL-30B at 59.8\%) with 3.3$\times$ fewer parameters, demonstrating the superior data efficiency unlocked by PKC synthesis and HaPO training.}
  \label{fig:comparison}
\end{figure*}

% ══════════════════════════════════════════════════
% 2. Related Work
% ══════════════════════════════════════════════════
\section{Related Work}

\noindent \textbf{Deep Research Agents.}
Augmenting LLMs with search tools and training them via reinforcement learning has become a dominant paradigm for building autonomous research agents.
On the text side, Search-o1~\cite{li2025searcho1} first integrated agentic search into reasoning chains by dynamically invoking search engines during inference.
Search-R1~\cite{jin2025searchr1} trained LLMs end-to-end with outcome-based RL to autonomously decide when and what to search.
ReSearch~\cite{chen2025research} explored RL-driven interleaved search and reasoning, while REDSearcher~\cite{chu2026redsearcher} achieved cost-efficient long-horizon agent training through graph-structured task synthesis and mid-training.
In the multimodal domain, MMSearch-R1~\cite{wu2026mmsearchr1} introduced the first end-to-end RL framework for multimodal search.
WebWatcher~\cite{geng2025webwatcher} reformulated text QA as visual QA via reverse image search.
DeepEyes~\cite{hong2025deepeyes} incentivized ``thinking with images'' by treating the model's native visual grounding as a tool, trained end-to-end with RL without cold-start SFT.
Vision-DeepResearch~\cite{huang2026visiondeepresearch} enabled MLLMs to perform tens of reasoning steps with hundreds of search engine interactions through cold-start SFT and RL training.
VSearcher~\cite{zhang2026vsearcher} transformed static multimodal models into long-horizon tool-calling search agents.
ASearcher~\cite{gao2025asearcher} scaled the interaction horizon beyond 100 turns through fully asynchronous RL.
OpenSearch-VL~\cite{chen2026opensearchvl} provided a fully open-source recipe combining Wikipedia-based multi-hop data synthesis with fatal-aware GRPO.

\noindent \textbf{RL Environment Scaling.}
The scarcity of high-quality interactive environments has become a central bottleneck for agentic RL training~\cite{chu2026agenticworldmodeling}.
In the general tool-use setting, ScaleEnv~\cite{tu2026scaleenv}, DreamGym~\cite{chen2026dreamgym}, and Agent-World~\cite{dong2026agentworld} addressed this through diverse environment synthesis, world-model-based experience generation, and self-evolving co-evolution of environments and policies, respectively.
CUA-Gym~\cite{wang2026cuagym} and Gym-Anything~\cite{aggarwal2026gymanything} further automated environment construction for computer-use and arbitrary software scenarios.
For search agents, the challenge extends to matched multi-hop reasoning data.
WebSailor-V2~\cite{li2025websailorv2} pioneered KG-based search data synthesis on a cyclic knowledge graph with a paired simulated environment.
WebDancer~\cite{wu2025webdancer} proposed scalable QA synthesis with two-stage SFT-then-RL training.
HopChain~\cite{wang2026hopchain} synthesized multi-hop vision-language data for RLVR.
On the environment side, SearchGym~\cite{zhang2026searchgym} and LiteResearcher~\cite{li2026literesearcher} constructed high-fidelity search simulators to reduce real-engine costs.
Agentic Proposing~\cite{jiao2026agenticproposing} and Socratic-Zero~\cite{wang2025socraticzero} explored agent-driven problem synthesis via goal-directed decision processes and multi-agent co-evolution.

\noindent \textbf{Agentic Reinforcement Learning.}
Agentic RL has become the standard paradigm for training strong agents, with recent flagship models such as Kimi K2.5~\cite{kimik25}, GLM-5~\cite{glm5}, Qwen3.5~\cite{qwen35}, and DeepSeek-V4~\cite{deepseekv4} all adopting it as a core post-training component.
The algorithmic foundation rests on GRPO~\cite{grpo}, which eliminates the value model via group-relative advantage estimation.
DAPO~\cite{dapo} scaled GRPO to industrial-level training, while SAPO~\cite{sapo} replaced hard clipping with a smooth sigmoid gate for improved long-sequence stability.
However, these methods assign a uniform trajectory-level advantage to all tokens, causing credit assignment to degrade in multi-turn agent tasks.
To address this, GiGPO~\cite{gigpo} introduced nested group structures for step-level credit assignment.
ARPO~\cite{arpo} combined entropy-guided agentic rollouts with step-level reward allocation for multi-turn web agents.
AT$^2$PO~\cite{at2po} unified turn-level tree search with policy optimization through entropy-guided expansion and turn-wise credit.
An alternative line tackles credit assignment through denser reward signals.
PRIME~\cite{prime} pioneered implicit process rewards that can be estimated online without step-level annotation.
More recently, Agent-RRM~\cite{agentrrm} extended this idea to agent trajectories with multi-dimensional structured critiques.

% ══════════════════════════════════════════════════
% 3. Method
% ══════════════════════════════════════════════════
\section{Method}

\subsection{Overview}
\label{sec:overview}
Given an image $I_0$ and a natural-language question $q$, a multimodal search agent interacts with a search environment $\mathcal{E}$ over multiple turns to retrieve evidence and reason toward an answer. We formalize this process as an MDP $(\mathcal{S}, \mathcal{A}, \mathcal{E}, R)$:
\begin{itemize}[leftmargin=1.5em, itemsep=2pt]
  \item \textbf{State.} $s_t = (I_0, q, z_1, a_1, o_1, \ldots, z_{t-1}, a_{t-1}, o_{t-1})$ is the full interaction history up to turn $t$.
  \item \textbf{Action.} Following the ReAct paradigm~\cite{yao2023react}, the agent generates a thought $z_t$ (natural-language reasoning) followed by an action $a_t$ (a structured tool call or a terminal answer). We write $y_t = (z_t, a_t)$ for the full generation at turn $t$.
  \item \textbf{Observation.} If $a_t$ is a tool call, the environment returns a textual observation $o_t = \mathcal{E}(a_t)$ (e.g., retrieved document snippets). If $a_t$ is a terminal answer, the episode ends.
  \item \textbf{Trajectory.} $\tau = (z_1, a_1, o_1, \ldots, z_T, a_T)$, where $a_T$ is the terminal answer.
  \item \textbf{Reward.} $R(\tau) \in \{0, 1\}$ indicates whether the final answer is correct.
  \item \textbf{Policy.} $\pi_\theta(y_t \mid s_t)$, parameterized by an MLLM, generates each turn $y_t = (z_t, a_t)$ autoregressively.
\end{itemize}
The training objective is $J(\theta) = \mathbb{E}[R(\tau)]$. During training, only agent-generated tokens (within $z_t$ and $a_t$) receive gradient; environment observation tokens $o_t$ are masked. SearchEyes uses a typed knowledge graph $\mathcal{G}$ as a unified backbone that simultaneously defines the search environment (\S3.2), drives training data synthesis (\S3.3), and provides step-level reward anchors for RL training (\S3.4).
\subsection{Knowledge Graph and Search Environment}
\label{sec:env}

\noindent \textbf{Typed Knowledge Graph.}
We construct a typed knowledge graph from three complementary sources.
\textbf{Wikidata5M}~\cite{wang2021kepler} provides structured relation triples $(h, p, t)$.
\textbf{Wiki6M} (OVEN-Wiki)~\cite{hu2023openvocabulary} provides entity-level text---title, summary, and full article.
\textbf{Wikipedia images} provide visual grounding for a subset of entities.
We define the resulting knowledge graph as $\mathcal{G} = (\mathcal{V}, \mathcal{R}, \mathcal{T}, \varphi)$, where $\mathcal{V}$ is the entity set, $\mathcal{R} \subseteq \mathcal{V} \times \mathcal{P} \times \mathcal{V}$ is the set of relation triples over predicate set $\mathcal{P}$, $\mathcal{T} = \{\textsc{Person}, \textsc{Work}, \textsc{Org}, \textsc{Geo}\}$ is a set of four semantic domains, and $\varphi\colon \mathcal{P} \to \mathcal{T}$ maps each predicate to its semantic domain.
An entity $v$ is retained in $\mathcal{V}$ only if it appears in all three sources, ensuring that every node simultaneously possesses structured relations, a textual description, and a visual depiction.
We further partition $\mathcal{V}$ into visual entities $\mathcal{V}_v$ (those whose Wikipedia image passes a quality filter) and knowledge-only entities $\mathcal{V}_k = \mathcal{V} \setminus \mathcal{V}_v$.
Hub entities with degree exceeding $d_{\max}$ and predicates on a manually curated blacklist (e.g., \texttt{instance\_of}, \texttt{subclass\_of}) are removed to avoid trivial shortcuts.

\noindent \textbf{Self-Contained Search Environment.}
For each entity $v \in \mathcal{V}$, let $\mathrm{doc}(v)$ denote its Wikipedia summary from Wiki6M.
The document collection $\mathcal{D} = \{\mathrm{doc}(v) \mid v \in \mathcal{V}\}$ forms a self-contained knowledge base, eliminating the need for external search APIs and ensuring fully deterministic, reproducible retrieval.
We implement the retrieval engine as a hybrid system combining two complementary signals:
\begin{itemize}[leftmargin=1.5em, itemsep=2pt]
  \item \textbf{Dense retrieval.} We precompute text embeddings $\mathbf{e}_v \in \mathbb{R}^d$ for each entity and store them as a memory-mapped matrix $\mathbf{E} \in \mathbb{R}^{|\mathcal{V}| \times d}$. Given a text query $q$, we compute $\mathrm{rank}_{\text{dense}}(q) = \mathrm{argsort}_{v}\; \cos(\mathbf{e}_q, \mathbf{e}_v)$.
  \item \textbf{Sparse retrieval.} We build a BM25 inverted index over entity titles and summaries, yielding $\mathrm{rank}_{\text{BM25}}(q)$.
\end{itemize}
The two rankings are fused via Reciprocal Rank Fusion (RRF):
\begin{equation}
  \mathrm{score}_{\text{RRF}}(q, v) = \sum_{r \in \{\text{dense},\, \text{BM25}\}} \frac{1}{\kappa + \mathrm{rank}_r(q, v)},
  \label{eq:rrf}
\end{equation}
where $\kappa$ is a smoothing constant.

\noindent \textbf{Tool Definitions.}
The agent is equipped with five tools. Three are deterministic retrieval tools operating directly on the search environment $\mathcal{E}$:
\begin{itemize}[leftmargin=1.5em, itemsep=2pt]
  \item $\texttt{text\_search}(q) \to \{(v_i, \mathrm{snippet}_i)\}_{i=1}^{k}$: returns the top-$k$ entities under $\mathrm{score}_{\text{RRF}}$, each accompanied by a snippet from $\mathrm{doc}(v_i)$.
  \item $\texttt{lookup}(v) \to \mathrm{doc}(v)$: returns the full document for entity $v$ identified by its Wikidata QID.
  \item $\texttt{visual\_search}(I, \mathrm{bbox}) \to \{(v_i, \mathrm{snippet}_i)\}_{i=1}^{k}$: crops the specified bounding box from image $I$, computes its image embedding, and retrieves the top-$k$ visually similar entities via cosine similarity against precomputed image embeddings $\mathbf{E}^{\text{img}} \in \mathbb{R}^{|\mathcal{V}_v| \times d}$.
\end{itemize}
The remaining two are auxiliary tools: $\texttt{summarize}(o, q)$ extracts query-relevant sentences from a long observation, and $\texttt{python\_interpreter}(c)$ executes sandboxed Python for lightweight computation. All observations are truncated to $L_{\text{obs}}$ tokens.
Since every document in $\mathcal{D}$ corresponds to exactly one entity via its Wikidata QID, a bijection $\mathrm{doc}(v) \leftrightarrow v$ is established.
The environment therefore records the exact entity accessed at each retrieval step during rollouts, enabling precise alignment between retrieved entities and the gold entity sequence $\mathbf{e}^* = (v_1, \ldots, v_K)$ produced by PKC synthesis (\S3.3), which provides step-level anchors for credit assignment in \S3.4.

\subsection{Perception-Knowledge Chain Synthesis}
\label{sec:pkc}

\begin{figure*}[tb!]
  \centering
  \includegraphics[width=\textwidth]{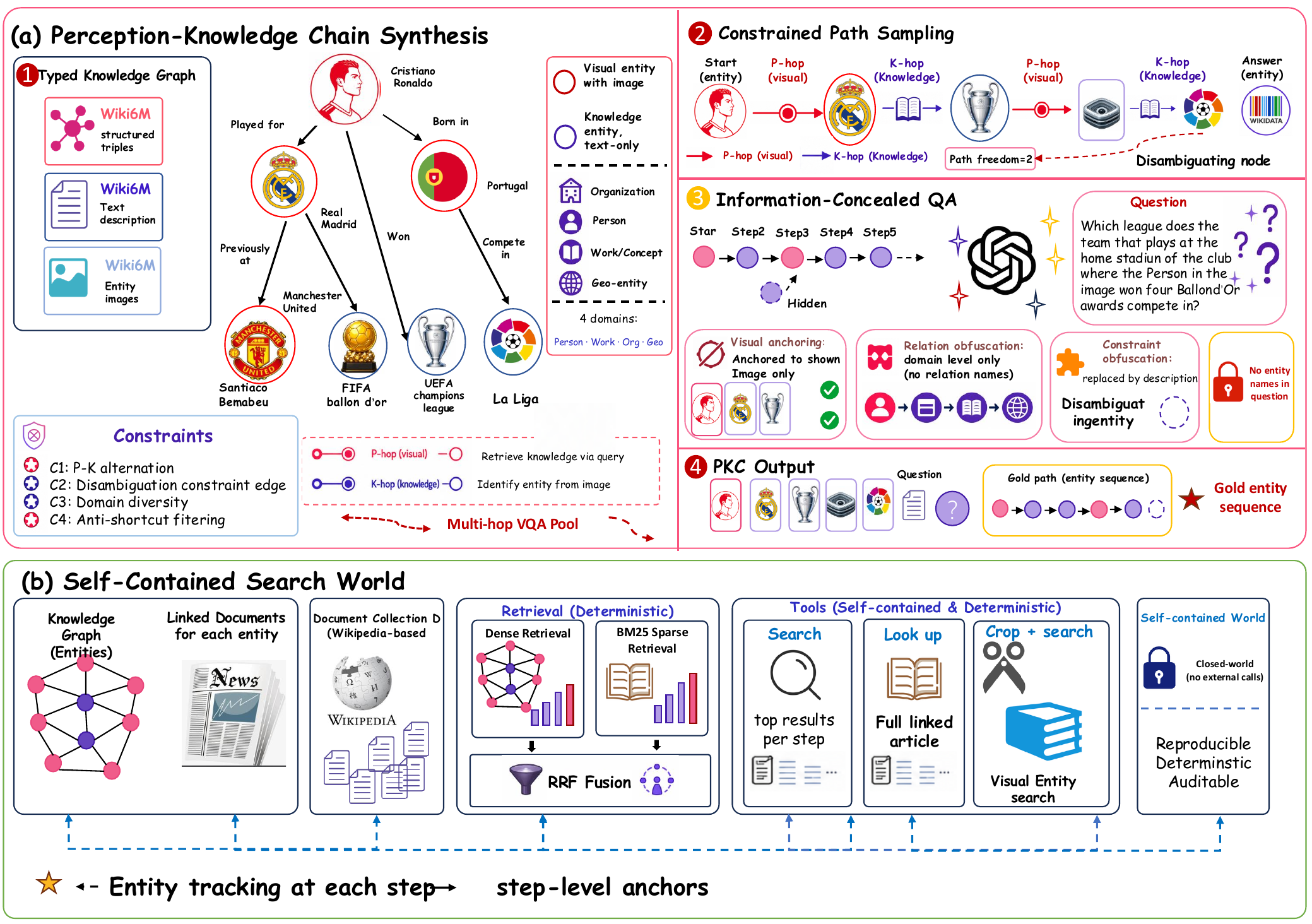}
  \caption{\textbf{Method architecture.} (a)~Perception-Knowledge Chain (PKC) synthesis: starting from a typed knowledge graph, constrained path sampling produces multi-hop questions with strict P--K alternation and disambiguating constraints (treewidth${\leq}$2). Information concealment ensures no entity name leaks into the question. (b)~Self-contained search world: the same knowledge graph defines a fully deterministic retrieval environment with hybrid BM25+dense retrieval fused via RRF, eliminating external API dependencies and enabling reproducible training.}
  \label{fig:method}
\end{figure*}

This section describes the pipeline from the typed knowledge graph $\mathcal{G}$ to supervised fine-tuning data (see Figure~\ref{fig:method} for an overview).
The pipeline proceeds in three stages: constrained path sampling on $\mathcal{G}$, information-concealed question generation, and trajectory synthesis in the search environment.

\noindent \textbf{Constrained Path Sampling.}
From $\mathcal{G}$, we sample multi-hop paths via a constrained random walk. Starting from an anchor entity $v_1 \in \mathcal{V}_v$ that possesses at least two outgoing edges, the walk produces a path of length $K$:
\begin{equation}
  \mathbf{p} = \bigl(v_1 \xrightarrow{r_1} v_2 \xrightarrow{r_2} \cdots \xrightarrow{r_K} v_{K+1}\bigr),
  \label{eq:path}
\end{equation}
where each $r_k \in \mathcal{P}$ is a predicate and $(v_k, r_k, v_{k+1}) \in \mathcal{R}$.
The walk is subject to the following constraints:

\smallskip
\noindent \emph{(C1) Perception-Knowledge alternation.}
Each hop $k$ is typed as either a perception hop (P) or a knowledge hop (K).
A P-hop targets a visual entity $v_{k+1} \in \mathcal{V}_v$ and requires the agent to identify it from an image; a K-hop targets a knowledge entity $v_{k+1} \in \mathcal{V}_k$ and requires text-based search.
A hop is typed P if it targets a visual entity ($v_{k+1} \in \mathcal{V}_v$) and the preceding hop was K; otherwise it is typed K. The first hop is always P (the anchor is visual). We enforce strict alternation (no consecutive P-hops) and require at least two P-hops per path, guaranteeing that each question exercises visual perception multiple times.

\noindent \emph{(C2) Disambiguating constraint.}
We classify each predicate as one-to-one (the relation from a parent yields a unique target) or one-to-few (multiple sibling targets exist).
At least one hop in $\mathbf{p}$ must use a one-to-few predicate, introducing answer ambiguity that cannot be resolved by following the main chain alone.
To resolve this ambiguity, we attach a disambiguating constraint edge to the answer entity $v_{K+1}$.
Let $\mathrm{Sib}(v_{K+1}) = \{u \neq v_{K+1} \mid (v_K, r_K, u) \in \mathcal{R}\}$ denote the set of sibling entities reachable from $v_K$ via the same predicate $r_K$.
We select a constraint predicate $r_c \in \mathcal{P}$ and a constraint entity $v_c \in \mathcal{V}$ such that the triple $(v_{K+1}, r_c, v_c)$ satisfies:
\begin{equation}
  (v_{K+1}, r_c, v_c) \in \mathcal{R} \quad \land \quad \forall\, u \in \mathrm{Sib}(v_{K+1}): (u, r_c, v_c) \notin \mathcal{R}.
  \label{eq:constraint}
\end{equation}
The constraint edge raises the treewidth of the reasoning graph from~1 to~2, so the agent must jointly satisfy the main-chain traversal and verify the constraint property.

\smallskip
\noindent \emph{(C3) Semantic domain diversity.}
Let $\mathbf{d} = \bigl(\varphi(r_1), \ldots, \varphi(r_K)\bigr) \in \mathcal{T}^K$ be the domain sequence of $\mathbf{p}$, where each $\varphi(r_k)$ maps predicate $r_k$ to one of the four semantic domains in $\mathcal{T}$.
The path must satisfy both local diversity and global coverage:
\begin{equation}
  \varphi(r_k) \neq \varphi(r_{k+1}) \;\;\forall\, k \in \{1, \ldots, K{-}1\}, \qquad |\{\varphi(r_k)\}_{k=1}^{K}| \geq 3.
  \label{eq:domain}
\end{equation}
The first condition prevents monotonic same-domain chains; the second ensures heterogeneous knowledge coverage.

\smallskip
\noindent \emph{(C4) Anti-shortcut filtering.}
To prevent trivially solvable paths, we apply three filtering rules during the walk:
(i)~hub exclusion: entities whose degree in $\mathcal{G}$ exceeds a threshold $d_{\max}$ are excluded, as high-degree nodes are easily guessable;
(ii)~predicate blacklist: meta-level predicates (e.g., ``instance of'', ``subclass of'') that do not constitute meaningful reasoning steps are rejected;
(iii)~deduplication: each candidate path is deduplicated by its anchor-answer pair $(v_1, v_{K+1})$ and by the full entity sequence to prevent near-duplicate questions.

\smallskip
Each sampled path retains the gold entity sequence $\mathbf{e}^* = (v_1, v_2, \ldots, v_{K+1})$ as metadata, which will serve as step-level anchors in~\S\ref{sec:hapo}.

\noindent \textbf{Information-Concealed Question Generation.}
Each path $\mathbf{p}$ is converted into a natural-language question $q$ by an LLM generator, subject to three concealment constraints that prevent the agent from bypassing the multi-hop reasoning chain:

\smallskip
\noindent \emph{(i) Visual anchoring.}
The anchor entity $v_1$ is presented solely through its image $I_0$; its name and all known aliases are withheld from $q$, forcing the first hop to require visual identification.

\smallskip
\noindent \emph{(ii) Relation obfuscation.}
The full predicate sequence $(r_1, \ldots, r_K)$ is never disclosed.
Instead, a single domain-level hint $\tilde{d}$ is provided, obtained by applying the domain mapping $\varphi$ to a randomly selected intermediate hop: $\tilde{d} = \varphi(r_j)$, $j \sim \mathrm{Uniform}\{2, \ldots, K{-}1\}$.
Since $\varphi$ is a many-to-one mapping from predicates to semantic domains, $\tilde{d}$ reveals the topic area but not the specific relation, preserving search ambiguity.

\smallskip
\noindent \emph{(iii) Constraint obfuscation.}
The disambiguating entity $v_c$ is never referred to by its exact name.
Instead, the generator replaces it with a partial or relational descriptor, requiring the agent to independently search for and verify $v_c$.

\smallskip
\noindent Collectively, the generated question must satisfy a non-leakage condition: no entity name or alias from the path $(v_1, \ldots, v_{K+1})$ may appear in the question text $q$.
Questions whose answers are trivially guessable (overly popular entities or single-token answers) are further filtered out.

\noindent \textbf{Trajectory Synthesis.}
The final stage generates expert search trajectories for supervised fine-tuning.
Since the synthesized questions are deliberately difficult (multi-hop, information-concealed), a generator policy $\pi_g$ achieves a low success rate under the unmodified environment $\mathcal{E}$.
We therefore introduce privileged generation conditions that simplify the task during synthesis; these conditions are stripped when exporting the final SFT data, so the trained policy faces the full, unmodified environment.

Given a question $q$ with image $I_0$ and gold entity sequence $\mathbf{e}^*$, the generator produces a trajectory $\tau = \bigl(z_1, a_1, o_1, \ldots, z_T, a_T\bigr)$, where each intermediate step consists of a reasoning trace $z_t$, a tool action $a_t$, and an observation $o_t = \widetilde{\mathcal{E}}(a_t)$, and $a_T$ is the terminal answer.
Three privileged mechanisms are applied during generation:

\smallskip
\noindent \emph{(i) Retrieval boost.}
The generator interacts with a modified environment $\widetilde{\mathcal{E}}$ in which the retrieval scores of gold entities are amplified.
For any search query $q'$ issued during rollout and any entity $v$ in the corpus, the boosted score is:
\begin{equation}
  \widetilde{\mathrm{score}}(q', v) =
  \begin{cases}
    \beta \cdot \mathrm{score}_{\text{RRF}}(q', v), & \text{if } v \in \mathbf{e}^*, \\
    \mathrm{score}_{\text{RRF}}(q', v), & \text{otherwise},
  \end{cases}
  \label{eq:boost}
\end{equation}
where $\beta > 1$ is the boost factor.
This increases the probability that gold-entity documents appear in top-ranked results without altering the generator's action distribution, as it simply encounters relevant documents more readily.

\smallskip
\noindent \emph{(ii) Observation denoising.}
Raw search observations $o_t$ returned by $\widetilde{\mathcal{E}}$ are often lengthy and noisy.
During generation, a summarizer produces a denoised observation $\tilde{o}_t$ that highlights query-relevant facts while discarding irrelevant content.
The generator conditions on $\tilde{o}_t$ (rather than $o_t$) to select its next action.

\smallskip
\noindent \emph{(iii) Rejection sampling.}
For each question $q$, we draw $N$ independent trajectories from $\pi_g$ under $\widetilde{\mathcal{E}}$.
A trajectory is accepted if and only if its final answer matches the gold answer (via normalized string matching).

\smallskip
\noindent \textbf{SFT training.}
The privileged conditions exist solely to increase the success rate of trajectory generation; the student policy never interacts with $\widetilde{\mathcal{E}}$.
Each accepted trajectory is exported as an offline sequence where the denoised observation $\tilde{o}_t$ is replaced with the raw observation $o_t$ originally returned by the environment, so that the student learns to reason over realistic, noisy retrieval results.
The exported sample takes the form $(q, I_0, z_1, a_1, o_1, \ldots, z_T, a_T)$, consistent with the trajectory definition in \S\ref{sec:overview}.
The student policy $\pi_\theta$ is trained via standard causal language modeling on the agent-generated tokens (thoughts $z_t$, actions $a_t$, and the final answer), with environment observations $o_t$ masked from the loss.
Let $\mathbf{x} = (x_1, \ldots, x_L)$ denote the full token sequence of an exported sample, and let $M_{\text{gen}}(\ell) \in \{0,1\}$ indicate whether token $x_\ell$ belongs to agent-generated spans (i.e., within some $z_t$, $a_t$, or the final answer). The SFT objective is:
\begin{equation}
  \mathcal{L}_{\text{SFT}} = -\sum_{\ell=1}^{L} M_{\text{gen}}(\ell) \cdot \log \pi_\theta(x_\ell \mid x_{<\ell}),
  \label{eq:sft}
\end{equation}
where observation tokens ($o_t$ spans) are provided as context but excluded from the loss.
Each sample additionally retains $\mathbf{e}^*$ as metadata for step-level credit assignment in~\S\ref{sec:hapo}.

\subsection{Hop-Anchored Policy Optimization}
\label{sec:hapo}

\begin{figure*}[tb!]
  \centering
  \includegraphics[width=\textwidth]{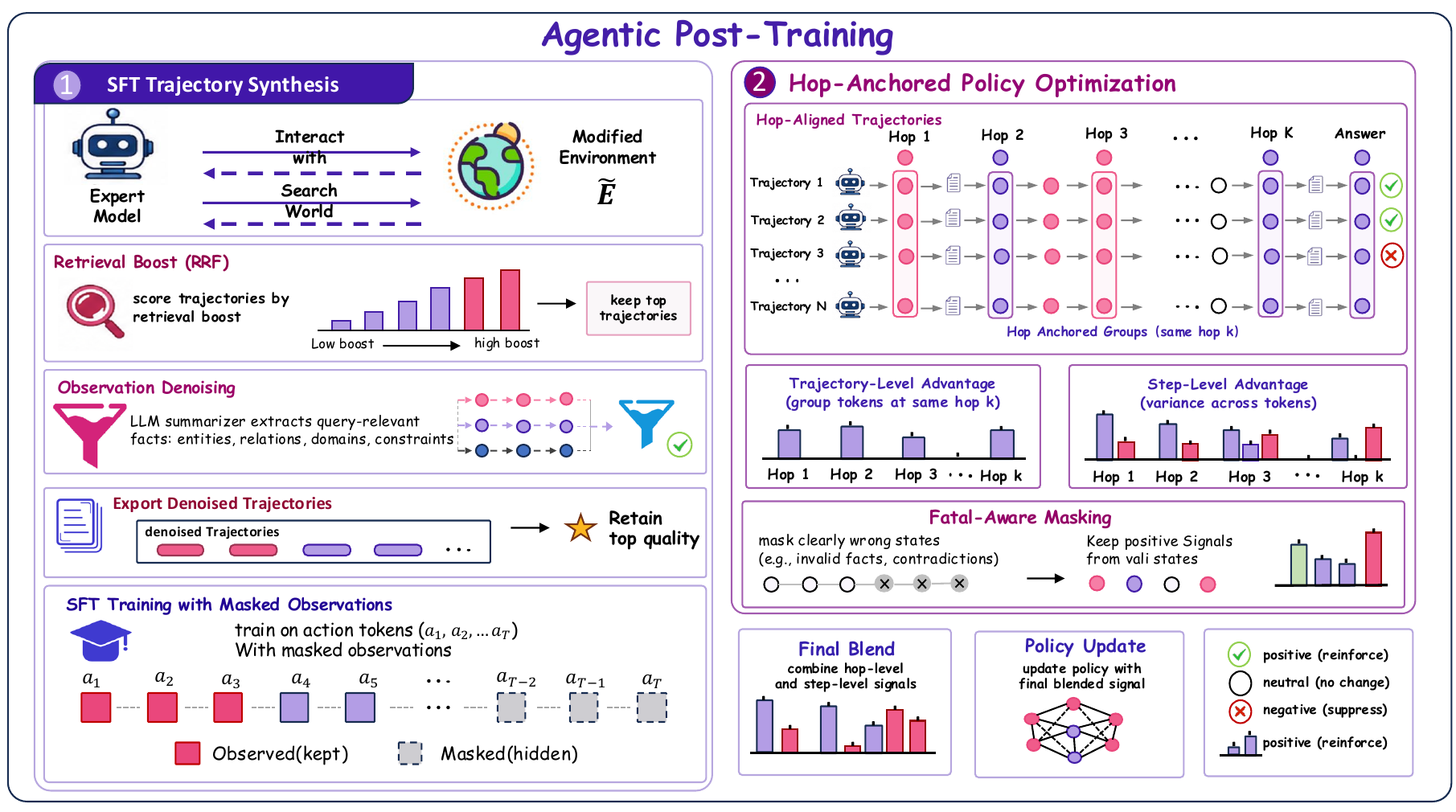}
  \caption{\textbf{Agentic post-training pipeline.} Stage~1 (SFT): an expert model generates trajectories in a retrieval-boosted environment; accepted trajectories are scored by retrieval boost, denoised via LLM summarization, and exported with raw observations for masked-observation supervised training. Stage~2 (HaPO): online rollouts are grouped by shared hop anchors to compute step-level advantages; fatal-aware masking suppresses degenerate suffixes while one-sided clamping retains positive signals from valid reasoning prefixes.}
  \label{fig:train}
\end{figure*}

Standard RL training for search agents relies on trajectory-level rewards: a binary signal indicating whether the final answer is correct.
For long-horizon multi-hop trajectories, the probability of a correct final answer decreases sharply with the number of hops, causing the trajectory-level reward to become extremely sparse and the resulting advantage estimates to carry high variance.
We address this by introducing \textbf{HaPO} (\textbf{H}op-\textbf{A}nchored \textbf{P}olicy \textbf{O}ptimization), which exploits the gold entity sequence $\mathbf{e}^*$ retained from PKC synthesis to construct step-level credit assignment signals without training a separate process reward model (Figure~\ref{fig:train}).

\noindent
HaPO builds on GRPO~\cite{grpo}.
For a question $q$, we sample $G$ trajectories $\{\tau^{(1)}, \ldots, \tau^{(G)}\}$ from the current policy $\pi_\theta$.
The standard GRPO objective is:
\begin{equation}
  \mathcal{J}_{\text{GRPO}} = \mathbb{E}_{q \sim \mathcal{Q}} \; \frac{1}{G} \sum_{i=1}^{G} \sum_{\ell=1}^{|\tau^{(i)}|} \min\!\Big(r_\ell^{(i)} \hat{A}^{(i)},\; \mathrm{clip}\big(r_\ell^{(i)}, 1{-}\epsilon, 1{+}\epsilon\big) \hat{A}^{(i)}\Big) - \lambda \, D_{\mathrm{KL}}(\pi_\theta \| \pi_{\text{ref}}),
  \label{eq:grpo}
\end{equation}
where $r_\ell^{(i)} = {\pi_\theta(x_\ell \mid x_{<\ell})} \big/ {\pi_{\text{old}}(x_\ell \mid x_{<\ell})}$ is the per-token importance ratio, $\pi_{\text{old}}$ is the policy snapshot at the start of the current update epoch, $\pi_{\text{ref}}$ is a fixed reference policy (the SFT checkpoint) used for the KL anchor, $\hat{A}^{(i)}$ is the trajectory-level advantage shared across all tokens in $\tau^{(i)}$, and $\lambda$ controls the KL penalty.
Since GRPO assigns a uniform advantage to every token within a trajectory, it provides no step-level differentiation for multi-hop reasoning.

\smallskip
\noindent \textbf{Trajectory-level advantage.}
Each trajectory receives a binary outcome reward $R^{(i)} \in \{0, 1\}$ based on final-answer correctness.
The trajectory-level advantage is:
\begin{equation}
  \hat{A}_{\text{ep}}^{(i)} = \frac{R^{(i)} - \bar{R}}{\sigma_R + \epsilon}, \qquad \bar{R} = \frac{1}{G}\sum_{j=1}^{G} R^{(j)}, \qquad \sigma_R = \mathrm{std}\!\big(\{R^{(j)}\}_{j=1}^{G}\big).
  \label{eq:episode_adv}
\end{equation}

\noindent \textbf{Hop-anchored step-level advantage.}
The key insight is that trajectories sharing a common intermediate state should be compared at that state to isolate the effect of subsequent actions.
In the search setting, exact state matching is infeasible due to the stochastic nature of retrieval.
Instead, we define a semantic anchor: trajectory $\tau^{(i)}$ is anchored at gold entity $v_k \in \mathbf{e}^*$ if $v_k$ appears in the retrieved observations at some step $t_k^{(i)}$, i.e., $v_k \in \mathrm{entities}(o_{t_k^{(i)}}^{(i)})$.
Intuitively, two trajectories that have both successfully retrieved the same gold entity have reached an equivalent intermediate state, regardless of the exact search queries or observation text that led them there.

For each gold entity $v_k$, we form a hop anchor group:
\begin{equation}
  \mathcal{H}_k = \bigl\{(i, t_k^{(i)}) \;\big|\; v_k \in \mathrm{entities}(o_{t_k^{(i)}}^{(i)}),\; t_k^{(i)} = \min\{t : v_k \in \mathrm{entities}(o_t^{(i)})\}\bigr\},
  \label{eq:hop_group}
\end{equation}
containing all trajectory-step pairs where entity $v_k$ was first retrieved.
Within each group $\mathcal{H}_k$ with $|\mathcal{H}_k| \geq 2$ and non-zero outcome variance, we compute group-relative advantages:
\begin{equation}
  \hat{A}_{\text{hop}}^{(i, t_k^{(i)})} = \frac{R^{(i)} - \bar{R}_k}{\sigma_{R_k} + \epsilon},
  \label{eq:hop_adv}
\end{equation}
where $\bar{R}_k$ and $\sigma_{R_k}$ are the mean and standard deviation of outcomes within $\mathcal{H}_k$.
This advantage is assigned at the anchor step $t_k^{(i)}$ and propagated to all subsequent tokens in trajectory $\tau^{(i)}$: for all $\ell$ corresponding to steps $t \geq t_k^{(i)}$, we set $\hat{A}_{\text{hop}}^{(i, \ell)} = \hat{A}_{\text{hop}}^{(i, t_k^{(i)})}$.
When a token falls under multiple anchor groups, the most recent (latest) anchor takes precedence, as it provides the most precise credit attribution.
For tokens not covered by any anchor group, we set $\hat{A}_{\text{hop}}^{(i, \ell)} = \hat{A}_{\text{ep}}^{(i)}$, falling back to the trajectory-level signal.

\noindent \textbf{Combined advantage.}
The final per-token advantage interpolates between the two levels:
\begin{equation}
  \hat{A}_{\text{final}}^{(i, \ell)} = \alpha \cdot \hat{A}_{\text{ep}}^{(i)} + (1 - \alpha) \cdot \hat{A}_{\text{hop}}^{(i, \ell)},
  \label{eq:hapo_combined}
\end{equation}
where $\alpha \in [0, 1]$ controls the mixing ratio.
When $\alpha = 1$, HaPO reduces to standard GRPO; when $\alpha = 0$, it relies entirely on hop-anchored signals.

\noindent \textbf{Fatal-aware masking.}
In long-horizon tool-use settings, consecutive tool errors (e.g., malformed queries, timeouts) can produce degenerate trajectory suffixes from which no useful gradient signal can be extracted.
We detect fatal steps: if a trajectory exhibits $M$ or more consecutive tool errors starting at step $t_f$, all tokens $\ell$ corresponding to steps $t \geq t_f$ are masked by setting $\hat{A}_{\text{final}}^{(i, \ell)} = 0$.
For the valid prefix ($t < t_f$), we apply one-sided clamping $\hat{A}_{\text{final}}^{(i, \ell)} \leftarrow \max(\hat{A}_{\text{final}}^{(i, \ell)}, 0)$, retaining positive signals while avoiding penalizing actions that preceded an exogenous tool failure.

\noindent \textbf{Policy optimization.}
We replace the hard clipping in Eq.~\eqref{eq:grpo} with a smooth sigmoid-based gate for stable long-horizon optimization.
The gating function uses asymmetric temperatures $\gamma^{+}$ and $\gamma^{-}$ (with $\gamma^{-} > \gamma^{+}$):
\begin{equation}
  g(r_\ell, \hat{A}) =
  \begin{cases}
    \sigma\!\bigl(\gamma^{+}(r_\ell - 1)\bigr) \cdot \tfrac{4}{\gamma^{+}}, & \text{if } \hat{A} > 0, \\[4pt]
    \sigma\!\bigl(\gamma^{-}(r_\ell - 1)\bigr) \cdot \tfrac{4}{\gamma^{-}}, & \text{if } \hat{A} \leq 0,
  \end{cases}
  \label{eq:gate}
\end{equation}
where $\sigma(\cdot)$ is the sigmoid function.
The larger $\gamma^{-}$ applies stronger damping when the advantage is negative, preventing large policy updates on undesirable actions.

The full HaPO objective applies the generation mask $M_{\text{gen}}(\ell)$ (defined identically to Eq.~\eqref{eq:sft}) to exclude observation tokens:
\begin{equation}
  \mathcal{L}_{\text{HaPO}} = -\frac{1}{G} \sum_{i=1}^{G} \sum_{\ell=1}^{|\tau^{(i)}|} M_{\text{gen}}(\ell) \cdot g\!\bigl(r_\ell^{(i)},\, \hat{A}_{\text{final}}^{(i,\ell)}\bigr) \cdot \hat{A}_{\text{final}}^{(i, \ell)} \;+\; \lambda \, D_{\mathrm{KL}}(\pi_\theta \| \pi_{\text{ref}}),
  \label{eq:hapo_loss}
\end{equation}
where $\lambda$ is the KL penalty coefficient.

% ======================================================
% 4. Experiments
% ======================================================
\section{Experiments}
\label{sec:experiments}

\subsection{Experimental Setup}
\label{sec:exp_setup}

\textbf{Knowledge base.}
We construct the entity knowledge graph by intersecting Wikidata5M~\cite{wang2021kepler} (4.6M entities, 822 relations, 20.6M triples), Wiki6M/OVEN~\cite{hu2023openvocabulary} (6.06M entity descriptions, 2.03M canonical images), and the associated Wikipedia entity images.
After intersecting the three sources and applying hub-exclusion and predicate-blacklist filtering, the final typed graph contains ${\sim}$1.2M entities (of which ${\sim}$340K are visual entities with quality-filtered images), connected by ${\sim}$5.8M relation triples over 822 predicates.

\textbf{Data synthesis.}
Using the PKC synthesizer (\S\ref{sec:pkc}), we sample multi-hop visual questions with graph-constrained P--K alternating structure and apply five filtering stages (degenerate-path, single-hop shortcut, knowledge-contamination, cross-modal grounding, and answer verifiability).
The filtered pool is split into two disjoint parts: 10K unique questions for SFT trajectory distillation and 12K questions for RL rollouts.
Through rejection sampling on the SFT questions (\S\ref{sec:pkc}), we collect 30K accepted expert trajectories (roughly three per question) for supervised fine-tuning.

\textbf{Base model.}
We use \textbf{Qwen3.5-9B}~\cite{qwen35} as our base vision--language model. As a native multimodal model with early-fusion vision--language pretraining, Qwen3.5-9B provides strong visual perception without requiring a separate vision encoder.

\textbf{Training.}
SFT: full-parameter fine-tuning on 8$\times$H20 GPUs for 3 epochs with learning rate $2{\times}10^{-5}$ and cosine schedule.
RL: HaPO (\S\ref{sec:hapo}) with mixing ratio $\alpha{=}0.3$, gate temperatures $\gamma^{+}{=}1.0$ and $\gamma^{-}{=}1.05$ (asymmetric, damping negative-advantage updates more strongly), fatal threshold $M{=}3$, group size $G{=}8$, and KL coefficient $\lambda{=}0.02$, trained for 200 optimization steps on 64 H20 GPUs.

\textbf{Tool environment.}
Our agent operates with five tools: \texttt{text\_search}, \texttt{visual\_search}, \texttt{lookup}, \texttt{summarize}, and \texttt{python\_interpreter}.
This tool set unifies knowledge retrieval (\texttt{text\_search}, \texttt{lookup}) with active visual perception (\texttt{visual\_search}) and auxiliary reasoning (\texttt{summarize}, \texttt{python\_interpreter}), all operating deterministically over the self-contained search world.

\textbf{Evaluation benchmarks.}
We evaluate on six established multimodal knowledge-intensive benchmarks:
\textbf{SimpleVQA}~\cite{cheng2025simplevqa} (factual visual QA),
\textbf{VDR}~\cite{zeng2026vdr} (visual deep research),
\textbf{MMSearch}~\cite{jiang2024mmsearch} (multimodal web search),
\textbf{LiveVQA}~\cite{fu2025livevqa} (real-time visual QA),
\textbf{BrowseComp-VL}~\cite{geng2025browsecompvl} (abbreviated BC-VL; hard multimodal browsing comprehension requiring long-horizon web navigation),
and \textbf{FVQA}~\cite{wang2017fvqa} (fact-based visual QA).
We additionally evaluate on our proposed \textbf{VisSearch Bench} (multi-hop visual search with P--K structure; see Appendix~\ref{app:vissearch}).
Correctness is judged by GPT-4o following the VDR evaluation protocol~\cite{zeng2026vdr}.

\textbf{Baselines.}
We compare against three categories of methods:
(1)~\emph{Direct Reasoning}: models answer from parametric knowledge without tools;
(2)~\emph{Agentic Workflow}: proprietary models equipped with a search toolset via prompting;
(3)~\emph{Multimodal DeepResearch MLLM}: open-source models specifically trained for iterative multimodal search and reasoning.

% -----------------------------------------------
% Main Results Table
% -----------------------------------------------
\subsection{Main Results}
\label{sec:main_results}

\begin{table*}[!tbp]
\centering
\caption{\textbf{Performance on multimodal knowledge-intensive QA and web-search benchmarks.}
\textbf{Bold} and \underline{underline} mark the best and second-best in each column.
BC-VL denotes BrowseComp-VL.
Avg.\ is computed over all six benchmarks; -- denotes unavailable scores.}
\label{tab:main}
\vspace{4pt}
\small
\renewcommand{\arraystretch}{1.15}
\resizebox{\textwidth}{!}{%
\begin{tabular}{@{}l cccccc c@{}}
\toprule
\textbf{Model} & \textbf{SimpleVQA} & \textbf{VDR} & \textbf{MMSearch} & \textbf{LiveVQA} & \textbf{BC-VL} & \textbf{FVQA} & \textbf{Avg.} \\
\midrule
\multicolumn{8}{c}{\cellcolor{cuhklightpink}\textbf{Direct Reasoning}} \\
\midrule
GPT-4o~\cite{team2024gpt4o}                    & 51.7 & 1.7  & 18.7 & 28.1 & 5.5  & 48.0 & 25.6 \\
GPT-5~\cite{openai2025gpt5}                    & 61.6 & 9.8  & 35.1 & 44.4 & 48.6 & 54.4 & 42.3 \\
Gemini-2.5-Flash~\cite{comanici2025geminiflash} & 57.9 & 6.2  & 30.4 & 51.0 & 37.1 & 47.7 & 38.4 \\
Gemini-2.5-Pro~\cite{comanici2025gemini}        & 63.0 & 8.0  & 39.8 & 60.3 & 43.1 & 60.7 & 45.8 \\
Claude-4-Sonnet~\cite{anthropic2025claude4}     & 50.9 & 2.0  & 18.7 & 38.5 & 29.3 & 35.3 & 29.1 \\
Claude-3.7-Sonnet~\cite{anthropic2025claude37}  & 42.7 & 4.6  & 21.1 & 38.0 & 32.3 & 36.7 & 29.2 \\
Qwen3-VL-8B~\cite{bai2025qwen3vl}              & 47.1 & 2.8  & 11.7 & 23.1 & 24.1 & 24.2 & 22.2 \\
Qwen3-VL-30B-A3B~\cite{bai2025qwen3vl}         & 53.2 & 3.8  & 18.7 & 42.7 & 29.6 & 34.7 & 30.5 \\
Qwen3-VL-32B~\cite{bai2025qwen3vl}             & 58.0 & 4.1  & 19.8 & 45.5 & 30.8 & 34.1 & 32.1 \\
\midrule
\multicolumn{8}{c}{\cellcolor{cuhklightpink}\textbf{Agentic Workflow}} \\
\midrule
GPT-5~\cite{openai2025gpt5}                    & 67.3 & 17.6 & 62.7 & --   & \underline{57.6} & 62.0 & -- \\
Gemini-2.5-Pro~\cite{comanici2025gemini}        & 74.3 & 10.0 & 55.7 & --   & 42.3 & 65.0 & -- \\
Gemini-3.1-Pro~\cite{google2026gemini31pro}     & --   & --   & \textbf{86.1} & \underline{76.6} & \textbf{64.1} & \textbf{84.0} & -- \\
Claude-4-Sonnet~\cite{anthropic2025claude4}     & --   & 13.6 & 67.2 & --   & 48.6 & 69.0 & -- \\
Claude-4.6-Opus~\cite{anthropic2026claudeopus}  & --   & --   & 76.2 & 67.4 & 48.3 & 74.5 & -- \\
Kimi-K2.5~\cite{kimik25}                       & --   & --   & 76.6 & 76.6 & 50.3 & 76.5 & -- \\
\midrule
\multicolumn{8}{c}{\cellcolor{cuhklightpink}\textbf{Multimodal DeepResearch Models}} \\
\midrule
Visual-ARFT-7B~\cite{liu2025visualarft}         & 42.4 & 3.3  & 34.5 & 25.4 & 16.5 & 41.7 & 27.3 \\
MMSearch-R1-7B~\cite{wu2025mmsearchr1}          & 57.4 & 2.9  & 53.8 & 48.4 & 20.9 & 58.4 & 40.3 \\
DeepEyes-V2-7B~\cite{hong2025deepeyes}          & 59.4 & 7.8  & 63.7 & --   & --   & 60.6 & -- \\
WebWatcher-7B~\cite{geng2025webwatcher}          & 54.3 & 10.3 & 49.1 & 51.2 & 21.2 & --   & -- \\
Qwen3-VL-8B~\cite{bai2025qwen3vl}              & 52.0 & 17.0 & 37.4 & 50.6 & 27.9 & 58.7 & 40.6 \\
SenseSearch-8B~\cite{chng2025sensesearch}        & 61.7 & 19.4 & 67.4 & 56.2 & 35.1 & 67.1 & 51.2 \\
ODE-8B~\cite{huang2026ode}                      & 70.3 & 20.4 & 66.0 & --   & 41.9 & 64.7 & -- \\
OpenSearch-VL-8B~\cite{feng2026opensearchvl}     & 71.6 & 20.8 & 64.5 & 59.6 & 37.6 & 71.5 & 54.3 \\
\cmidrule{1-8}
Qwen3-VL-30B~\cite{bai2025qwen3vl}             & 55.1 & 20.2 & 44.2 & 62.0 & 34.1 & 63.0 & 46.4 \\
ODE-30B~\cite{huang2026ode}                     & 71.0 & 26.4 & 69.7 & --   & 46.1 & 69.3 & -- \\
OpenSearch-VL-30B~\cite{feng2026opensearchvl}    & 74.9 & 33.5 & 68.7 & 67.4 & 41.1 & 73.2 & 59.8 \\
\cmidrule{1-8}
WebWatcher-32B~\cite{geng2025webwatcher}         & 59.0 & --   & 55.3 & 58.7 & 27.0 & --   & -- \\
OpenSearch-VL-32B~\cite{feng2026opensearchvl}    & \underline{76.2} & \underline{33.8} & 72.3 & 70.5 & 43.8 & 74.7 & \underline{61.9} \\
\midrule
\multicolumn{8}{c}{\cellcolor{cuhklightpurple}\textbf{SearchEyes (Ours)}} \\
\midrule
Qwen3.5-9B (Agentic)~\cite{qwen35}             & 56.8 & 19.3 & 41.6 & 55.4 & 31.7 & 63.2 & 44.7 \\
\rowcolor{cuhklightpurple}
\textbf{SearchEyes-9B (Ours)}                   & 75.4 & 28.3 & 69.2 & 66.1 & 42.3 & 74.2 & 59.3 \\
\cmidrule{1-8}
Qwen3.5-27B (Agentic)~\cite{qwen35}            & 64.3 & 25.8 & 59.1 & 50.6 & 39.4 & 67.3 & 51.1 \\
\rowcolor{cuhklightpurple}
\textbf{SearchEyes-27B (Ours)}                  & \textbf{80.9} & \textbf{39.4} & \underline{82.4} & \textbf{77.3} & 49.3 & \underline{79.1} & \textbf{68.1} \\
\bottomrule
\end{tabular}}
\end{table*}

Table~\ref{tab:main} presents our main results across six multimodal knowledge-intensive benchmarks.
Figure~\ref{fig:comparison} provides a visual comparison against closed-source models (left) and illustrates the parameter efficiency of SearchEyes relative to open-source baselines (right).
We highlight the following observations:

\textbf{(1) Direct reasoning is insufficient for knowledge-intensive visual queries.}
Even frontier models like GPT-5 and Gemini-2.5-Pro achieve modest scores on search-demanding benchmarks such as VDR (${\leq}$10\%) and MMSearch (${\leq}$40\%) in direct-reasoning mode.
Our base model Qwen3.5-9B with a generic agentic workflow reaches 44.7 average accuracy, yet still falls well short of the trained SearchEyes-9B (59.3), confirming that effective tool use requires dedicated training beyond prompting alone.

\textbf{(2) Tool access alone does not unlock search capability.}
Equipping models with an agent workflow improves over direct reasoning, but open-source models still substantially underperform proprietary systems. Qwen3-VL-30B with agent prompting reaches only 44.2 on MMSearch---well below the 86.1 achieved by Gemini-3.1-Pro with the same tool access---indicating that effective tool use requires dedicated training.

\textbf{(3) SearchEyes achieves competitive performance at 9B scale.}
Despite being trained on a single 9B base model, SearchEyes achieves strong results across all benchmarks, benefiting from the structured PKC data synthesis and hop-aware RL optimization.
On VisSearch Bench (Appendix~\ref{app:vissearch}), which explicitly requires multi-hop P--K alternating reasoning, SearchEyes demonstrates particular strength, validating that our training recipe instills genuine compositional search behavior rather than superficial tool-call patterns.

% -----------------------------------------------
% Ablation Studies
% -----------------------------------------------
\subsection{Ablation Study}
\label{sec:ablation}

We conduct comprehensive ablations on the two core contributions of SearchEyes: the PKC data synthesis pipeline and the HaPO training algorithm. All ablations use Qwen3.5-9B as the base model and report accuracy on four benchmarks (SimpleVQA, VDR, MMSearch, FVQA) as well as the average.

\subsubsection{PKC Data Quality Ablations}

\begin{table}[!htbp]
\centering
\small
\caption{\textbf{PKC ablation.} Each structural constraint contributes to final performance. We report per-benchmark accuracy and the average over four benchmarks.}
\label{tab:pkc_ablation}
\vspace{2pt}
\begin{tabular}{@{}lccccc@{}}
\toprule
\textbf{Configuration} & \textbf{SVQA} & \textbf{VDR} & \textbf{MMS} & \textbf{FVQA} & \textbf{Avg.} \\
\midrule
Full PKC pipeline                       & 75.4 & 28.3 & 69.2 & 74.2 & \textbf{61.8} \\
\quad w/o P--K alternation constraint     & 72.1 & 22.6 & 64.8 & 70.8 & 57.6 \\
\quad w/o treewidth${\leq}$2 constraint   & 73.8 & 25.1 & 66.4 & 72.0 & 59.3 \\
\quad w/o anti-shortcut filtering         & 69.5 & 20.8 & 59.7 & 66.3 & 54.1 \\
\quad w/o domain diversity balancing      & 74.2 & 26.0 & 67.1 & 72.5 & 60.0 \\
\quad w/o cross-modal grounding filter    & 72.6 & 24.7 & 65.8 & 71.0 & 58.5 \\
\quad w/o information concealment         & 73.0 & 25.4 & 66.9 & 71.8 & 59.3 \\
\quad w/o retrieval boost ($\beta{=}1$)   & 71.8 & 23.2 & 63.5 & 69.7 & 57.1 \\
\midrule
GPT-4o free-form synthesis (no graph)    & 65.2 & 16.4 & 52.3 & 62.8 & 49.2 \\
Wikipedia path sampling~\cite{feng2026opensearchvl} & 68.7 & 19.2 & 58.6 & 67.4 & 53.5 \\
HopChain-style augmentation              & 70.1 & 21.5 & 61.0 & 68.9 & 55.4 \\
\bottomrule
\end{tabular}
\end{table}

Table~\ref{tab:pkc_ablation} validates each design choice in our PKC pipeline.
Removing the P--K alternation constraint produces questions that collapse into single-modality chains (e.g., all-text retrieval), reducing the agent's learned multi-modal switching behavior and causing the largest single-constraint drop ($-4.2$ avg).
The anti-shortcut filtering is equally critical ($-7.7$ avg): without it, the agent learns to exploit single-hop retrieval patterns rather than genuine multi-step reasoning, with particularly severe degradation on VDR ($-7.5$) where multi-hop knowledge linking is essential.
The treewidth constraint prevents degenerate graph structures that lead to trivially decomposable questions ($-2.5$ avg), while cross-modal grounding ($-3.3$ avg) ensures that visual entities are meaningfully connected to the reasoning chain.
Information concealment prevents entity name leakage into the question text, and removing it ($-2.5$ avg) shows the agent learns shallow matching instead of genuine visual identification.
Retrieval boost ($\beta$) applied during SFT data selection ensures training trajectories are non-trivial; setting $\beta{=}1$ (no boost) degrades by $-4.7$ avg as easy examples dominate the training mix.
Domain diversity balancing has a moderate effect ($-1.8$ avg), primarily affecting generalization across diverse query types.
The HopChain-style augmentation ($-6.4$ avg) confirms that naive chain-extension heuristics do not substitute for graph-constrained structural guarantees.
Compared to the free-form GPT-4o synthesis baseline ($-12.6$ avg) and Wikipedia path sampling used in OpenSearch-VL~\cite{feng2026opensearchvl} ($-8.3$ avg), our graph-constrained pipeline with structural guarantees yields substantially higher-quality training data.

\subsubsection{HaPO Algorithm Ablations}

\begin{table}[!htbp]
\centering
\small
\caption{\textbf{HaPO ablation.} Hop-anchored credit assignment provides the largest gain over standard GRPO. We report per-benchmark accuracy and $\Delta$ relative to Standard GRPO.}
\label{tab:hapo_ablation}
\vspace{2pt}
\begin{tabular}{@{}lccccc c@{}}
\toprule
\textbf{Configuration} & \textbf{SVQA} & \textbf{VDR} & \textbf{MMS} & \textbf{FVQA} & \textbf{Avg.} & $\bm{\Delta}$ \\
\midrule
\textbf{Full HaPO (ours)}                & 75.4 & 28.3 & 69.2 & 74.2 & \textbf{61.8} & \textbf{+4.0} \\
\quad w/o hop-anchored advantage          & 73.1 & 25.8 & 66.0 & 71.5 & 59.1 & +1.3 \\
\quad w/o smooth asymmetric gating        & 74.2 & 27.0 & 67.8 & 72.8 & 60.5 & +2.7 \\
\quad w/o fatal-aware masking             & 73.8 & 26.5 & 67.2 & 72.2 & 59.9 & +2.1 \\
\quad w/o one-sided advantage clamping    & 74.5 & 27.3 & 68.0 & 73.0 & 60.7 & +2.9 \\
\quad w/o observation token masking       & 74.0 & 26.8 & 67.5 & 72.5 & 60.2 & +2.4 \\
\quad w/ fixed $\alpha{=}1$ (ep-only)     & 73.4 & 25.5 & 65.8 & 71.2 & 59.0 & +1.2 \\
\quad w/ fixed $\alpha{=}0$ (hop-only)    & 74.8 & 27.5 & 68.5 & 73.5 & 61.1 & +3.3 \\
\midrule
Standard GRPO~\cite{grpo}        & 72.0 & 24.5 & 64.3 & 70.2 & 57.8 & 0.0 \\
GiGPO~\cite{gigpo}              & 72.8 & 25.2 & 65.1 & 70.8 & 58.5 & +0.7 \\
DAPO~\cite{dapo}                   & 72.3 & 24.8 & 64.8 & 70.5 & 58.1 & +0.3 \\
\bottomrule
\end{tabular}
\end{table}

Table~\ref{tab:hapo_ablation} isolates the contribution of each HaPO component.
The hop-anchored advantage provides the largest individual gain ($+4.0$ vs.\ $+1.3$ without it), confirming that decomposing credit assignment by reasoning hops is far more effective than trajectory-level rewards for multi-step search tasks.
The impact is most pronounced on VDR ($28.3 \to 25.8$) and MMSearch ($69.2 \to 66.0$), which require the longest reasoning chains.
Fatal-aware masking contributes $+2.1$ by preventing gradient noise from cascading tool failures that otherwise corrupt policy updates in long trajectories.
One-sided advantage clamping ($+2.9$) preserves useful signals from valid reasoning prefixes that precede exogenous failures, while our smooth asymmetric gating ($+2.7$) replaces hard clipping with a sigmoid-based gate that prevents large updates on undesirable actions.
Excluding observation tokens from the loss ($+2.4$) prevents the policy from memorizing retrieval noise.
Sensitivity to the mixing ratio $\alpha$ shows that pure hop-only signal ($\alpha{=}0$, $+3.3$) outperforms pure trajectory-level ($\alpha{=}1$, $+1.2$), while the interpolated version ($\alpha{=}0.3$, $+4.0$) achieves the best balance.
Compared to standard GRPO ($+0.0$), GiGPO ($+0.7$), and DAPO ($+0.3$), HaPO demonstrates that existing RL improvements provide marginal gains for multi-hop search---the key bottleneck is step-level credit assignment, which HaPO directly addresses.

% -----------------------------------------------
% Qualitative Analysis
% -----------------------------------------------
\subsection{Qualitative Analysis}
\label{sec:qualitative}

Appendix~\ref{app:examples} presents a detailed 4-hop trajectory from SearchEyes.
The agent begins with a visual perception hop (P-hop): it identifies a museum building from the input image using \texttt{visual\_search}, noting that the museum network has look-alike branches designed by different architects.
This is followed by a knowledge retrieval hop (K-hop): the agent looks up the candidate branch's architect.
The trajectory continues with a second P-hop (verifying the city from the image background to rule out the distractor branch) and a final K-hop (searching for that architect's most prestigious award), culminating in the correct final answer.

This example highlights two key behaviors instilled by our training recipe:
(1)~the agent actively uses \texttt{visual\_search} to extract and disambiguate visual entities before committing to an answer, rather than guessing entity names;
(2)~the P--K alternation emerges naturally from HaPO's hop-anchored credit, which rewards each modality-switching decision independently.

% Flush all experiment-section floats into the body so tables never migrate
% into the appendix. Placed after all section text, so no paragraph is split.
\FloatBarrier

% ══════════════════════════════════════════════════
% 5. Conclusion
% ══════════════════════════════════════════════════
\section{Conclusion}

We presented \textbf{SearchEyes}, a data--algorithm co-designed framework for training multi-hop visual search agents. Our PKC synthesis pipeline generates high-quality training data with verifiable structural guarantees, while HaPO enables efficient step-level credit assignment by reusing the knowledge chain metadata as semantic anchors. Experiments demonstrate state-of-the-art performance with fewer search steps than competing methods.

% ── References ───────────────────────────────────
\bibliographystyle{unsrtnat}
\bibliography{references}

@inproceedings{li2025searcho1,
  title={Search-o1: Agentic Search-Enhanced Large Reasoning Models},
  author={Li, Xiaoxi and Dong, Guanting and Jin, Jiajie and Zhang, Yuyao and Zhou, Yujia and Zhu, Yutao and Zhang, Peitian and Dou, Zhicheng},
  booktitle={Proceedings of the Conference on Empirical Methods in Natural Language Processing (EMNLP)},
  year={2025}
}

@article{jin2025searchr1,
  title={Search-R1: Training LLMs to Reason and Leverage Search Engines with Reinforcement Learning},
  author={Jin, Bowen and Zeng, Hansi and Yue, Zhenrui and Wang, Dong and Han, Jiawei},
  journal={arXiv preprint arXiv:2503.09516},
  year={2025}
}

@inproceedings{chen2025research,
  title={ReSearch: Learning to Reason with Search for LLMs via Reinforcement Learning},
  author={Chen, Mingyang and Sun, Linzhuang and Li, Tianpeng and Sun, Haoze and Zhou, Yijie and Zhu, Chenzheng and Lu, Keer and Liang, Hao and Zhang, Wentao and Chen, Weipeng},
  booktitle={Advances in Neural Information Processing Systems (NeurIPS)},
  year={2025}
}

@article{chu2026redsearcher,
  title={REDSearcher: A Scalable and Cost-Efficient Framework for Long-Horizon Search Agents},
  author={Chu, Zheng and Wang, Xiao and Hong, Jack and Fan, Huiming and Huang, Yuqi and Yang, Yue and Xu, Guohai and Zhao, Chenxiao and Xiang, Cheng and Hu, Shengchao},
  journal={arXiv preprint arXiv:2602.14234},
  year={2026}
}

@inproceedings{wu2026mmsearchr1,
  title={MMSearch-R1: Incentivizing LMMs to Search},
  author={Wu, Jinming and Deng, Zihao and Li, Wei and Liu, Yiding and You, Bo and Li, Bo and Ma, Zejun and Liu, Ziwei},
  booktitle={Proceedings of the Association for Computational Linguistics (ACL)},
  year={2026}
}

@article{geng2025webwatcher,
  title={WebWatcher: Breaking New Frontier of Vision-Language Deep Research Agent},
  author={Geng, Xinyu and Xia, Peng and Zhang, Zhen and Wang, Xinyu and Wang, Qiuchen and Ding, Rui and Wang, Chao and Wu, Jialong and Zhao, Yida},
  journal={arXiv preprint arXiv:2508.05748},
  year={2025}
}

@article{huang2026visiondeepresearch,
  title={Vision-DeepResearch: Incentivizing DeepResearch Capability in Multimodal Large Language Models},
  author={Huang, Wenxuan and Zeng, Yu and Wang, Qiuchen and Fang, Zhen},
  journal={arXiv preprint arXiv:2601.22060},
  year={2026}
}

@article{zhang2026vsearcher,
  title={VSearcher: Long-Horizon Multimodal Search Agent via Reinforcement Learning},
  author={Zhang, Ruiyang and Sun, Qianguo and Song, Chao and Qi, Yiyan and Zheng, Zhedong},
  journal={arXiv preprint arXiv:2603.02795},
  year={2026}
}

@inproceedings{gao2025asearcher,
  title={Beyond Ten Turns: Unlocking Long-Horizon Agentic Search with Large-Scale Asynchronous RL},
  author={Gao, Jiaxuan and Fu, Wei and Xie, Minyang and Xu, Shusheng and He, Chuyi and Mei, Zhiyu and Zhu, Banghua and Wu, Yi},
  booktitle={Advances in Neural Information Processing Systems (NeurIPS)},
  year={2025},
  note={arXiv:2508.07976}
}

@article{chen2026opensearchvl,
  title={OpenSearch-VL: An Open Recipe for Frontier Multimodal Search Agents},
  author={Chen, Shuang and Feng, Kaituo and Chen, Hangting and Huang, Wenxuan and Dai, Dasen and Shou, Quanxin and Lin, Yunlong and Yue, Xiangyu and Pang, Tianyu},
  journal={arXiv preprint arXiv:2605.05185},
  year={2026}
}

@article{tu2026scaleenv,
  title={ScaleEnv: Scaling Environment Synthesis from Scratch for Generalist Interactive Tool-Use Agent Training},
  author={Tu, Dunwei and Hao, Hongyan and Yang, Hansi and Chen, Yihao and Zhang, Yi-Kai},
  journal={arXiv preprint arXiv:2602.06820},
  year={2026},
  note={ICML 2026}
}

@article{chen2026dreamgym,
  title={Scaling Agent Learning via Experience Synthesis},
  author={Chen, Zhaorun and Zhao, Zhuokai and Zhang, Kai and Liu, Bo and Qi, Qi},
  journal={arXiv preprint arXiv:2511.03773},
  year={2026},
  note={ICLR 2026}
}

@article{dong2026agentworld,
  title={Agent-World: Scaling Real-World Environment Synthesis for Evolving General Agent Intelligence},
  author={Dong, Guanting and Lu, Junting and Huang, Junjie and Zhong, Wanjun and Liu, Longxiang},
  journal={arXiv preprint arXiv:2604.18292},
  year={2026}
}

@article{wang2026cuagym,
  title={CUA-Gym: Scaling Verifiable Training Environments and Tasks for Computer-Use Agents},
  author={Wang, Bowen and Lu, Dunjie and Wang, Junli and Bai, Tianyi and Liu, Shixuan},
  journal={arXiv preprint arXiv:2605.25624},
  year={2026}
}

@article{aggarwal2026gymanything,
  title={Gym-Anything: Turn any Software into an Agent Environment},
  author={Aggarwal, Pranjal and Neubig, Graham and Welleck, Sean},
  journal={arXiv preprint arXiv:2604.06126},
  year={2026}
}

@article{li2025websailorv2,
  title={WebSailor-V2: Bridging the Chasm to Proprietary Agents via Synthetic Data and Scalable Reinforcement Learning},
  author={Li, Kuan and Zhang, Zhongwang and Yin, Huifeng and Ye, Rui and Zhao, Yida},
  journal={arXiv preprint arXiv:2509.13305},
  year={2025}
}

@article{wu2025webdancer,
  title={WebDancer: Towards Autonomous Information Seeking Agency},
  author={Wu, Jialong and Li, Baixuan and Fang, Runnan and Yin, Wenbiao and Zhang, Liwen},
  journal={arXiv preprint arXiv:2505.22648},
  year={2025},
  note={NeurIPS 2025}
}

@article{wang2026hopchain,
  title={HopChain: Multi-Hop Data Synthesis for Generalizable Vision-Language Reasoning},
  author={Wang, Shenzhi and Liu, Shixuan and Zhou, Jing and Gao, Chang and Chen, Xiong-Hui},
  journal={arXiv preprint arXiv:2603.17024},
  year={2026}
}

@article{zhang2026searchgym,
  title={SearchGym: Bootstrapping Real-World Search Agents via Cost-Effective and High-Fidelity Environment Simulation},
  author={Zhang, Xichen and He, Ziyi and Zhu, Yinghao and Wu, Sitong and Yu, Shaozuo},
  journal={arXiv preprint arXiv:2601.14615},
  year={2026}
}

@article{li2026literesearcher,
  title={LiteResearcher: A Scalable Agentic RL Training Framework for Deep Research Agent},
  author={Li, Wanli and Qu, Bince and Pan, Bo and Zhang, Jianyu and Liu, Zheng},
  journal={arXiv preprint arXiv:2604.17931},
  year={2026}
}

@article{jiao2026agenticproposing,
  title={Agentic Proposing: Enhancing Large Language Model Reasoning via Compositional Skill Synthesis},
  author={Jiao, Zhengbo and Wang, Shaobo and Zhang, Zifan and Ren, Xuan and Wang, Wei},
  journal={arXiv preprint arXiv:2602.03279},
  year={2026},
  note={ICML 2026}
}

@article{wang2025socraticzero,
  title={Socratic-Zero: Bootstrapping Reasoning via Data-Free Agent Co-evolution},
  author={Wang, Shaobo and Jiao, Zhengbo and Zhang, Zifan and Peng, Yilang and Ze, Xu},
  journal={arXiv preprint arXiv:2509.24726},
  year={2025}
}

@article{chu2026agenticworldmodeling,
  title={Agentic World Modeling: Foundations, Capabilities, Laws, and Beyond},
  author={Chu, Meng and Zhang, Xuan Billy and Lin, Kevin Qinghong and Kong, Lingdong and Zhang, Jize},
  journal={arXiv preprint arXiv:2604.22748},
  year={2026}
}

@article{kimik25,
  title={Kimi K2.5: Visual Agentic Intelligence},
  author={Kimi Team},
  journal={arXiv preprint arXiv:2602.02276},
  year={2026},
  note={Moonshot AI}
}

@misc{qwen35,
  title={Qwen3.5: Towards Native Multimodal Agents},
  author={Qwen Team},
  year={2026},
  howpublished={\url{https://qwen.ai/blog?id=qwen3.5}},
  note={Alibaba Cloud, Technical Report}
}

@article{glm5,
  title={GLM-5: from Vibe Coding to Agentic Engineering},
  author={Zhipu AI},
  journal={arXiv preprint arXiv:2602.15763},
  year={2026}
}

@misc{deepseekv4,
  title={DeepSeek-V4: Towards Highly Efficient Million-Token Context Intelligence},
  author={DeepSeek-AI},
  year={2026},
  howpublished={\url{https://huggingface.co/deepseek-ai/DeepSeek-V4-Pro}},
  note={Technical Report}
}

@article{grpo,
  title={DeepSeekMath: Pushing the Limits of Mathematical Reasoning in Open Language Models},
  author={Shao, Zhihong and Wang, Peiyi and Zhu, Qihao and Xu, Runxin and Song, Junxiao and Li, Xiao and Zhang, Haowei and Zhang, Mingchuan and Li, Y. K. and Wu, Y. and Guo, Daya},
  journal={arXiv preprint arXiv:2402.03300},
  year={2024},
  note={NeurIPS 2024; introduces GRPO}
}

@article{dapo,
  title={DAPO: An Open-Source LLM Reinforcement Learning System at Scale},
  author={Yu, Qiying and Zhang, Zheng and others},
  journal={arXiv preprint arXiv:2503.14476},
  year={2025},
  note={NeurIPS 2025; ByteDance}
}

@article{sapo,
  title={Soft Adaptive Policy Optimization},
  author={Gao, Jiajun and others},
  journal={arXiv preprint arXiv:2511.20347},
  year={2025},
  note={Qwen Team, Alibaba}
}

@article{gigpo,
  title={Group-in-Group Policy Optimization for LLM Agent Training},
  author={Feng, Lang and He, Shuo and others},
  journal={arXiv preprint arXiv:2505.10978},
  year={2025},
  note={NeurIPS 2025; step-level credit assignment via nested groups}
}

@article{arpo,
  title={Agentic Reinforced Policy Optimization},
  author={Dong, Zihan and Mao, Xingjian and others},
  journal={arXiv preprint arXiv:2507.19849},
  year={2025},
  note={RUC-NLPIR; entropy-guided agentic rollout + step-wise credit}
}

@article{at2po,
  title={{AT$^2$PO}: Agentic Turn-based Policy Optimization via Tree Search},
  author={Zong, Zefang and Chen, Dingwei and others},
  journal={arXiv preprint arXiv:2601.04767},
  year={2026},
  note={ICML 2026 Spotlight; turn-level tree exploration + credit assignment}
}

@article{prime,
  title={Process Reinforcement through Implicit Rewards},
  author={Cui, Ganqu and Yuan, Lifan and others},
  journal={arXiv preprint arXiv:2502.01456},
  year={2025},
  note={Online implicit PRM without step-level annotations}
}

@article{agentrrm,
  title={Exploring Reasoning Reward Model for Agents},
  author={Fan, Kexiang and Feng, Lang and others},
  journal={arXiv preprint arXiv:2601.22154},
  year={2026},
  note={S-Lab NTU / Shanghai AI Lab (MMLab); multi-faceted reasoning reward}
}

@article{li2025webthinker,
  title={WebThinker: Empowering Large Reasoning Models with Deep Research Capability},
  author={Li, Xiaoxi and Jin, Jiajie and Dong, Guanting and Qian, Hongjin and Zhu, Yutao and Wu, Yongkang and Wen, Ji-Rong and Dou, Zhicheng},
  journal={arXiv preprint arXiv:2504.21776},
  year={2025},
  note={NeurIPS 2025}
}

@article{deepdive2025,
  title={DeepDive: Synthesizing Hard-to-Find Questions for Deep Research},
  author={Zhang, Yuchen and others},
  journal={arXiv preprint arXiv:2509.10446},
  year={2025}
}

@article{wang2021kepler,
  title={KEPLER: A Unified Model for Knowledge Embedding and Pre-trained Language Representation},
  author={Wang, Xiaozhi and Gao, Tianyu and Zhu, Zhaocheng and Feng, Zhengyan and Liu, Zhiyuan and Li, Juanzi and Tang, Jian},
  journal={Transactions of the Association for Computational Linguistics},
  volume={9},
  pages={176--194},
  year={2021}
}

@inproceedings{hu2023openvocabulary,
  title={Open-domain Visual Entity Recognition: Towards Recognizing Millions of Wikipedia Entities},
  author={Hu, Hexiang and Yang, Yi and Luo, Jiacheng and Pang, Liangliang and Sigal, Leonid},
  booktitle={ICCV},
  year={2023}
}

@article{yao2023react,
  title={ReAct: Synergizing Reasoning and Acting in Language Models},
  author={Yao, Shunyu and Zhao, Jeffrey and Yu, Dian and Du, Nan and Shafran, Izhak and Narasimhan, Karthik and Cao, Yuan},
  journal={arXiv preprint arXiv:2210.03629},
  year={2023}
}

@article{team2024gpt4o,
  title={GPT-4o System Card},
  author={{OpenAI}},
  journal={OpenAI Technical Report},
  year={2024},
  url={https://cdn.openai.com/gpt-4o-system-card.pdf}
}

@article{openai2025gpt5,
  title={GPT-5 System Card},
  author={{OpenAI}},
  journal={OpenAI Technical Report},
  year={2025},
  url={https://openai.com/index/gpt-5-system-card/}
}

@article{comanici2025gemini,
  title={Gemini 2.5: Pushing the Frontier with Advanced Reasoning, Multimodality, Long Context, and Next Generation Agentic Capabilities},
  author={Comanici, Gheorghe and Bieber, Eric and Schaekermann, Mike and Pasupat, Panupong and Sachdeva, Noveen and others},
  journal={arXiv preprint arXiv:2507.06261},
  year={2025}
}

@article{anthropic2025claude4,
  title={Claude 4 Sonnet System Card},
  author={{Anthropic}},
  journal={Anthropic Technical Report},
  year={2025},
  url={https://www.anthropic.com/research/claude-4-sonnet-system-card}
}

@article{comanici2025geminiflash,
  title={Gemini 2.5 Flash: A Fast Yet Capable Multimodal Model},
  author={Comanici, Gheorghe and others},
  journal={Google DeepMind Technical Report},
  year={2025}
}

@article{anthropic2025claude37,
  title={The Claude Model Family: Claude 3.7 Sonnet},
  author={{Anthropic}},
  journal={Anthropic Model Card},
  year={2025},
  url={https://docs.anthropic.com/en/docs/about-claude/models}
}

@article{google2026gemini31pro,
  title={Gemini 3.1 Pro Technical Report},
  author={{Google DeepMind}},
  journal={Google DeepMind Technical Report},
  year={2026}
}

@article{anthropic2026claudeopus,
  title={Claude Opus 4 System Card},
  author={{Anthropic}},
  journal={Anthropic Technical Report},
  year={2026},
  url={https://www.anthropic.com/research/claude-opus-4-system-card}
}

@article{bai2025qwen3vl,
  title={Qwen3-VL Technical Report},
  author={Bai, Shuai and Cai, Yuxuan and Chen, Rui and Chen, Keqin and Chen, Xingyu and others},
  journal={arXiv preprint arXiv:2511.21631},
  year={2025}
}

@inproceedings{cheng2025simplevqa,
  title={SimpleVQA: Multimodal Factuality Evaluation for Multimodal Large Language Models},
  author={Cheng, Xiao and Zhang, Wei and Zhang, Shuai and Yang, Jie and Guan, Xinyu and Wu, Xingyu and others},
  booktitle={Proceedings of the IEEE/CVF International Conference on Computer Vision (ICCV)},
  pages={4637--4646},
  year={2025}
}

@article{zeng2026vdr,
  title={VDR-Bench: A Benchmark for Visual Deep Research},
  author={Zeng, Hansi and Huang, Shijue and Guo, Hangyu and Jin, Bowen and Han, Jiawei},
  journal={arXiv preprint arXiv:2603.11804},
  year={2026}
}

@inproceedings{jiang2024mmsearch,
  title={MMSearch: Benchmarking the Potential of Large Models as Multi-Modal Search Engines},
  author={Jiang, Dongyang and Ren, Jianqi and Lin, Wenqi and Lu, Yuxuan and Guo, Ziqi and Zhou, Yining},
  booktitle={Advances in Neural Information Processing Systems (NeurIPS)},
  year={2024}
}

@article{fu2025livevqa,
  title={LiveVQA: A Benchmark for Real-Time Visual Question Answering},
  author={Fu, Chaoyou and Dai, Yingjie and Luo, Yao and others},
  journal={arXiv preprint arXiv:2503.12527},
  year={2025}
}

@article{geng2025browsecompvl,
  title={BrowseComp-VL: Benchmarking Multimodal Web Browsing Comprehension},
  author={Geng, Xinyu and Xia, Peng and Wang, Xinyu and others},
  journal={arXiv preprint arXiv:2504.03957},
  year={2025}
}

@article{wang2017fvqa,
  title={FVQA: Fact-Based Visual Question Answering Requiring Multi-Hop Reasoning},
  author={Wang, Peng and Wu, Qi and Shen, Chunhua and Dick, Anthony and van den Hengel, Anton},
  journal={IEEE Transactions on Pattern Analysis and Machine Intelligence},
  volume={40},
  pages={2413--2427},
  year={2017}
}

@article{liu2025visualarft,
  title={Visual-ARFT: Reinforcement Fine-Tuning for Visual Agentic Search},
  author={Liu, Wei and Zhang, Yi and Li, Hao and others},
  journal={arXiv preprint arXiv:2505.08964},
  year={2025}
}

@article{wu2025mmsearchr1,
  title={MMSearch-R1: Incentivizing Multimodal LLMs to Search},
  author={Wu, Jinming and Deng, Zihao and Li, Wei and Liu, Yiding and You, Bo and Li, Bo and Ma, Zejun and Liu, Ziwei},
  journal={arXiv preprint arXiv:2502.02038},
  year={2025}
}

@article{hong2025deepeyes,
  title={DeepEyes-V2: Multi-Turn Visual Agentic Search with Progressive Visual Reasoning},
  author={Hong, Jack and Fan, Huiming and Huang, Yuqi and others},
  journal={arXiv preprint arXiv:2505.10428},
  year={2025}
}

@article{chng2025sensesearch,
  title={SenseSearch: Empowering Vision-Language Models with High-Resolution Agentic Search via Reinforcement Learning},
  author={Chng, Yi Xuan and Hu, Tao and Tong, Wei and Li, Xuan and Chen, Jiazhi and others},
  journal={arXiv preprint arXiv:2512.24330},
  year={2025}
}

@article{huang2026ode,
  title={Towards On-Policy Data Evolution for Visual-Native Multimodal Deep Search Agents},
  author={Huang, Shijue and Guo, Hangyu and Dong, Guanting and Li, Chenxin and Lu, Junting and Geng, Xinyu and Su, Zhaochen and Li, Zhenyu and Chen, Shuang and Wang, Hongru and Fung, Yi R.},
  journal={arXiv preprint arXiv:2605.10832},
  year={2026}
}

@article{feng2026opensearchvl,
  title={OpenSearch-VL: An Open Recipe for Frontier Multimodal Search Agents},
  author={Feng, Kaituo and Zhang, Minghan and Chen, Shuang and Lin, Yilan and Fan, Kefan and Jiang, Yingchun and Li, Haonan and Zheng, Danyang and Wang, Chenxu and Yue, Xiang},
  journal={arXiv preprint arXiv:2605.05185},
  year={2026}
}

% ── Appendix ─────────────────────────────────────
% ══════════════════════════════════════════════════
% Appendix
% ══════════════════════════════════════════════════
\newpage
\beginappendix

\section{VisSearch Bench}
\label{app:vissearch}

We introduce \textbf{VisSearch Bench}, a dedicated evaluation benchmark for multi-hop visual search that directly validates the quality of our PKC synthesis pipeline.
VisSearch Bench comprises 1000 questions over 935 unique Wikipedia entity images, with guaranteed P--K alternating structure, disambiguating constraints, and multi-domain coverage, making it uniquely suited for evaluating agents' compositional search capabilities.

\textbf{Construction.}
Questions are sampled from the same PKC pipeline described in \S\ref{sec:pkc}, subject to additional difficulty filters: minimum 4 hops, at least 2 perception hops requiring visual identification, and constraint edges that demand independent verification.
All questions are held out from training data.
Images are sourced from the Wiki6M dataset (OVEN-Wiki subset)~\cite{hu2023openvocabulary}, with each image corresponding to a Wikipedia entity used as a visual anchor.

\textbf{Data Format.}
Each benchmark entry contains: question ID, relative path to the anchor entity image, multi-hop question, gold answer (entity title), reasoning chain (ordered list of \texttt{from\_qid} $\to$ \texttt{relation} $\to$ \texttt{to\_qid} hops), constraint entities for validation, graph structure type (linear/branched), and number of hops.

\textbf{Evaluation Metric.}
We report \textbf{Exact Match (EM)} accuracy: a prediction is correct if its normalized answer matches the gold entity title after removing articles, punctuation, and case.

\textbf{Results.}
Table~\ref{tab:vissearch} reports performance on VisSearch Bench. SearchEyes demonstrates particular strength on this benchmark, confirming that the P--K alternating structure learned during training transfers to held-out evaluation.
Notably, even strong proprietary models (GPT-5, Kimi-K2.5) achieve $<$10\% accuracy, reflecting the genuine multi-hop difficulty that single-turn reasoning cannot bypass.

\begin{table}[tb!]
\centering
\small
\caption{\textbf{VisSearch Bench results.} Performance on our multi-hop visual search benchmark with P--K alternating structure.}
\label{tab:vissearch}
\vspace{2pt}
\begin{tabular}{@{}lc@{}}
\toprule
\textbf{Model} & \textbf{VisSearch Acc.} \\
\midrule
Qwen3.5-9B (Agentic)          & 3.8 \\
SearchEyes-9B                  & 18.6 \\
\cmidrule{1-2}
Qwen3.5-27B (Agentic)         & 5.1 \\
SearchEyes-27B                 & \textbf{24.3} \\
\cmidrule{1-2}
OpenSearch-VL-8B~\cite{feng2026opensearchvl}               & 6.8 \\
OpenSearch-VL-32B~\cite{feng2026opensearchvl}              & 9.4 \\
SenseSearch-8B~\cite{chng2025sensesearch}                 & 5.6 \\
\cmidrule{1-2}
GPT-5~\cite{openai2025gpt5}                          & 5.4 \\
Kimi-K2.5~\cite{kimik25}                      & 7.2 \\
\bottomrule
\end{tabular}
\end{table}

\section{PKC Synthesis Details}
\label{app:pkc}

\noindent\textbf{Knowledge Graph Statistics.}
The graph is built from Wikidata5M~\cite{wang2021kepler} (4.6M entities, 822 relations, 20.6M triples) intersected with Wiki6M/OVEN~\cite{hu2023openvocabulary} (6.06M entity descriptions, 2.03M images). After intersection and filtering, the typed knowledge graph contains 1.2M entities (of which 340K are visual entities with quality-filtered images), 5.8M relation triples, and 822 distinct predicates mapped to four semantic domains.

\noindent\textbf{Path Sampling Parameters.}
We sample paths of length $K \in \{3, 4, 5\}$ with distribution $P(K{=}3) = 0.3$, $P(K{=}4) = 0.5$, $P(K{=}5) = 0.2$.
The hub exclusion threshold is $d_{\max} = 500$.
The predicate blacklist contains 47 meta-level predicates.
Each path requires $\geq 2$ P-hops and covers $\geq 3$ semantic domains.

\noindent\textbf{Question Generation.}
We use GPT-4o as the question generator with temperature 0.7 and top-$p$ 0.95.
Each path produces one question candidate; questions whose answers appear in the top-100 most common Wikipedia entities are filtered out.
After deduplication and filtering, the pool contains 22K questions, split into 10K for SFT trajectory distillation and 12K for RL rollouts.

\noindent\textbf{Trajectory Synthesis.}
The retrieval boost factor is $\beta = 3.0$.
We draw $N = 8$ trajectories per question from the generator policy (GPT-4o) over the 10K SFT questions and accept those with correct final answers (acceptance rate $\sim$42\%).
Observation denoising uses a separate GPT-4o-mini call with the prompt: ``\textit{Extract only the facts relevant to the query from this document. Return a concise summary.}''
After denoising and export, the final SFT dataset contains $\sim$30K expert trajectories (roughly three per question).

\section{Training Hyperparameters}
\label{app:training}

\noindent\textbf{SFT Stage.}
Base model: Qwen3.5-9B.
Learning rate: $2 \times 10^{-5}$ with cosine schedule and 100-step warmup.
Batch size: 32 (effective, with gradient accumulation).
Max sequence length: 8192 tokens.
Training: 3 epochs over the 30K trajectory dataset.
Observation tokens are masked from the loss ($M_{\text{gen}}$ mask).

\noindent\textbf{HaPO Stage.}
Group size: $G = 8$ rollouts per question.
Mixing ratio: $\alpha = 0.3$ (favoring hop-anchored signals).
KL penalty: $\lambda = 0.02$.
Gate temperatures: $\gamma^{+} = 1.0$, $\gamma^{-} = 1.05$ (asymmetric; negative-advantage updates decay faster).
Fatal step threshold: $M = 3$ consecutive tool errors.
Learning rate: $5 \times 10^{-7}$ with linear warmup (50 steps) and constant schedule.
Training: 200 update steps over the 12K question pool with online rollouts.
Max trajectory length: 12 hops (tool calls).
Retrieval top-$k$: 5 documents per search call, each observation truncated to $L_{\text{obs}} = 4000$ characters.

\section{Agent System Prompt}
\label{app:prompt}

The SearchEyes agent uses a ReAct-style system prompt during both inference and RL rollouts. The full prompt is shown below.

\begin{tcolorbox}[
  colback=white, colframe=black!50,
  title={\sffamily\small\bfseries System Prompt (Evaluation)},
  fonttitle=\small\color{black},
  boxrule=0.5pt, arc=1pt, left=6pt, right=6pt, top=3pt, bottom=3pt
]
\small\ttfamily
You are a visual research assistant. Given an image and a complex multi-hop question, you must find the answer by searching through a knowledge base step by step.

\smallskip
When you have gathered sufficient information and are ready to provide the definitive response, you must enclose the entire final answer within <answer></answer> tags. The answer inside <answer> tags should be a short, precise entity name --- NOT a long sentence.

\smallskip
\textnormal{\textbf{Tools:}}
\smallskip

\{"type": "function", "function": \{"name": "text\_search", "description": "Search the knowledge base with a text query. Returns top results with titles, IDs, and summaries.", "parameters": \{"type": "object", "properties": \{"query": \{"type": "string"\}\}, "required": ["query"]\}\}\}

\smallskip
\{"type": "function", "function": \{"name": "visual\_search", "description": "Crop a region of the input image and retrieve visually similar entities from the knowledge base.", "parameters": \{"type": "object", "properties": \{"region": \{"type": "string"\}\}, "required": ["region"]\}\}\}

\smallskip
\{"type": "function", "function": \{"name": "lookup", "description": "Read the full Wikipedia content of a specific entity by its ID.", "parameters": \{"type": "object", "properties": \{"entity\_id": \{"type": "string"\}\}, "required": ["entity\_id"]\}\}\}

\smallskip
\textnormal{\textbf{Strategy:}}
\begin{enumerate}[leftmargin=1.5em]\small
\item First, carefully examine the image to identify the entity shown.
\item Use text\_search or visual\_search to find relevant entities and verify your identification.
\item Follow the chain of relationships described in the question step by step.
\item At each hop, search for or read the relevant entity and verify the connection before moving on.
\item Only provide your final answer when you have traced the COMPLETE chain and have sufficient evidence.
\item Your final answer must be a precise entity name enclosed in <answer></answer> tags.
\end{enumerate}
\end{tcolorbox}

For brevity, the box above lists the three core retrieval tools; the full prompt also exposes \texttt{summarize} and \texttt{python\_interpreter}, for a total of five tools (Appendix~\ref{app:tools}). During RL training, the same prompt structure is used but tool calls are handled natively by the \texttt{verl-agent} multi-turn rollout engine with \texttt{<tool\_call>} / \texttt{<tool\_response>} XML formatting and explicit Thought/Action formatting.

\section{Tool Specifications}
\label{app:tools}

Table~\ref{tab:tools} summarizes the tool set available to the SearchEyes agent. During RL training, tools operate against the self-contained PKC knowledge base (deterministic retrieval). During evaluation on external benchmarks, tools are backed by either the same KB (for VisSearch Bench) or a web search API (for SimpleVQA, MMSearch, etc.).

\begin{table}[tb!]
\centering
\small
\caption{\textbf{SearchEyes tool specifications.} All tools return text observations truncated to $L_{\text{obs}} = 4000$ characters.}
\label{tab:tools}
\vspace{2pt}
\begin{tabular}{@{}llp{5.5cm}@{}}
\toprule
\textbf{Tool} & \textbf{Input} & \textbf{Description} \\
\midrule
\texttt{text\_search} & query: str & Hybrid BM25+dense retrieval over Wiki6M KB. Returns top-$k$ entities with title, ID, and snippet. \\
\texttt{visual\_search} & image, region & Crop specified region (full/center/top\_half/etc.), compute image embedding, retrieve visually similar entities via cosine similarity. \\
\texttt{lookup} & entity\_id: str & Read full Wikipedia document of an entity by its Wikidata QID. Returns title, summary, and body text. \\
\texttt{summarize} & text, query & Extract query-relevant sentences from a long document using keyword overlap scoring. \\
\texttt{python\_interpreter} & code: str & Execute sandboxed Python for computation or string processing. Restricted to safe builtins only. \\
\bottomrule
\end{tabular}
\end{table}

\section{Evaluation Protocol}
\label{app:eval}

\noindent\textbf{Inference Setup.}
We deploy SearchEyes using vLLM with tensor parallelism across 8 H20 GPUs. The agent operates as a multi-turn ReAct loop: at each turn, it generates a thought and a tool call, receives the tool observation, and continues until producing a final \texttt{<answer>} tag or reaching the maximum turn limit. Inference uses temperature $T{=}0.6$, top-$p{=}0.95$, and a maximum of 50 turns per episode.

\noindent\textbf{Tool Backend.}
For external benchmarks (SimpleVQA, VDR, MMSearch, FVQA, LiveVQA, BrowseComp-VL), the \texttt{text\_search} tool is backed by the Serper web search API, and \texttt{visual\_search} uses an image-based web search service. For VisSearch Bench, all tools operate on the self-contained PKC knowledge base to ensure controlled, reproducible evaluation.

\noindent\textbf{Answer Extraction.}
The agent's final answer is extracted from the last occurrence of \texttt{<answer>...</answer>} tags in the response. If no such tag is found, we fall back to: (1)~\texttt{answer(text="...")} function-call format, (2)~``Answer: ...'' line format, (3)~the last non-empty line of the response.

\noindent\textbf{Scoring.}
Following the VDR evaluation protocol~\cite{zeng2026vdr}, we use an LLM-as-judge (GPT-4o) to assess answer correctness. The judge receives the question, gold answer, and model prediction, and outputs a binary correct/incorrect judgment. For VisSearch Bench, we additionally report two-tier string matching:
\begin{itemize}[leftmargin=1.5em]
\item \textbf{Exact Match (EM):} Normalized prediction equals normalized gold answer. Normalization removes articles (a/an/the), punctuation, and extra whitespace, then lowercases.
\item \textbf{Substring Match:} Normalized gold answer is contained within normalized prediction, or vice versa. Used as a secondary metric.
\end{itemize}

\noindent\textbf{Format Penalty.}
During RL training, if the model produces $>$10 \texttt{<answer>} tags (degenerate repetition), the reward is divided by 4.

\noindent\textbf{Repetition Detection.}
We monitor $n$-gram distinct ratio (window size 30 characters). If the ratio drops below 0.5 for 3 consecutive turns, the episode is terminated and the model is forced to produce a final answer.

\noindent\textbf{Baseline Reproduction.}
For baseline models that do not report results on all benchmarks, we run inference using their released checkpoints with the same tool backend and evaluation pipeline for fair comparison. For proprietary models (GPT-5, Gemini-2.5-Pro, etc.), we use their native agentic capabilities or API-based tool calling with the same evaluation prompts.

\section{Case Study: Multi-hop Agent Trajectory}
\label{app:examples}

We present a complete 4-hop SearchEyes trajectory to illustrate how the P--K alternating pattern emerges during inference. Each step shows the agent's internal reasoning (\textit{Thought}), tool invocation (\textit{Action}), and retrieved evidence (\textit{Observation}). The gold entity anchors used by HaPO for step-level credit assignment are annotated on the right.

\begin{tcolorbox}[
  colback=white, colframe=cuhkpurple, 
  title={\sffamily\bfseries\color{white} Case Study: 4-hop P--K Alternating Trajectory},
  fonttitle=\small,
  boxrule=0.8pt, arc=2pt, left=6pt, right=6pt, top=4pt, bottom=4pt
]

\small
\textbf{Question:} \textit{The image shows a titanium-clad museum that belongs to an international art-museum network whose branches were designed by different architects. Identify this specific branch, determine its architect, confirm the city using the scenery in the background, and name the most prestigious architecture prize that architect has received.}

\textbf{Gold Answer:} Pritzker Architecture Prize \hfill \textbf{Hops:} 4 (P$\to$K$\to$P$\to$K)

\textbf{Entity Chain:} $e_1$\,=\,Guggenheim Museum Bilbao $\to$ $e_2$\,=\,Frank Gehry $\to$ $e_3$\,=\,Bilbao \textit{(constraint)} $\to$ $e_4$\,=\,Pritzker Architecture Prize
\end{tcolorbox}

\begin{tcolorbox}[
  colback=cuhklightpurple, colframe=cuhkpurple!60,
  title={\sffamily\small\bfseries Step 1: Visual Identification (P-hop) \hfill Anchor: $e_1$},
  fonttitle=\small\color{cuhkdeeppurple},
  boxrule=0.5pt, arc=1pt, left=6pt, right=6pt, top=3pt, bottom=3pt
]
\small
\textbf{Thought:} The image shows a distinctive titanium-clad building beside a river --- this looks like a Guggenheim Museum, but the network has several branches by different architects. Let me identify it visually.

\textbf{Action:} \texttt{visual\_search(image, region="center")}

\textbf{Observation:} (1) \underline{Guggenheim Museum Bilbao} --- ``titanium-clad museum of modern art\ldots'' (2) \underline{Guggenheim Museum, New York} --- ``spiral building designed by Frank Lloyd Wright\ldots'' \textit{(a look-alike branch by a different architect --- the city must be confirmed).}
\end{tcolorbox}

\begin{tcolorbox}[
  colback=white, colframe=cuhkgold!80,
  title={\sffamily\small\bfseries Step 2: Knowledge Retrieval (K-hop) \hfill Anchor: $e_2$},
  fonttitle=\small\color{black},
  boxrule=0.5pt, arc=1pt, left=6pt, right=6pt, top=3pt, bottom=3pt
]
\small
\textbf{Thought:} The top match is the Bilbao branch. Let me look up its architect, while keeping in mind the New York branch (Frank Lloyd Wright) as a distractor to rule out.

\textbf{Action:} \texttt{lookup("Guggenheim Museum Bilbao")}

\textbf{Observation:} \underline{Guggenheim Museum Bilbao} --- ``designed by Canadian-American architect Frank Gehry; opened in 1997; located in Bilbao, Spain, on the Nerv\'{i}on River\ldots''
\end{tcolorbox}

\begin{tcolorbox}[
  colback=cuhklightpurple, colframe=cuhkpurple!60,
  title={\sffamily\small\bfseries Step 3: Visual Verification (P-hop) \hfill Anchor: $e_3$},
  fonttitle=\small\color{cuhkdeeppurple},
  boxrule=0.5pt, arc=1pt, left=6pt, right=6pt, top=3pt, bottom=3pt
]
\small
\textbf{Thought:} To rule out the New York branch (a different architect), I verify the city from the background, which shows a river and green hills.

\textbf{Action:} \texttt{visual\_search(image, region="background")}

\textbf{Observation:} (1) \underline{Bilbao} --- ``a city in the Basque Country of northern Spain on the Nerv\'{i}on estuary, surrounded by green mountains\ldots'' \textit{Confirms the Bilbao branch $\Rightarrow$ architect is Frank Gehry, not Frank Lloyd Wright.}
\end{tcolorbox}

\begin{tcolorbox}[
  colback=white, colframe=cuhkgold!80,
  title={\sffamily\small\bfseries Step 4: Final Knowledge Retrieval (K-hop) \hfill Anchor: $e_4$},
  fonttitle=\small\color{black},
  boxrule=0.5pt, arc=1pt, left=6pt, right=6pt, top=3pt, bottom=3pt
]
\small
\textbf{Thought:} The architect is confirmed as Frank Gehry. Now I search for the most prestigious architecture prize he has received.

\textbf{Action:} \texttt{text\_search("Frank Gehry most prestigious architecture prize")}

\textbf{Observation:} \underline{Frank Gehry} --- ``\ldots received the Pritzker Architecture Prize in 1989, widely regarded as architecture's most prestigious award\ldots''
\end{tcolorbox}

\begin{tcolorbox}[
  colback=cuhklightpurple!30, colframe=cuhkpurple,
  boxrule=0.8pt, arc=2pt, left=6pt, right=6pt, top=3pt, bottom=3pt
]
\small
\textbf{Final Answer:} Pritzker Architecture Prize \checkmark
\end{tcolorbox}

\smallskip
\noindent\textbf{Analysis.}
This trajectory demonstrates: (1)~the agent identifies the visual entity via \texttt{visual\_search} without any text hint; (2)~it resolves a genuine ambiguity---the Guggenheim network has branches by different architects (Bilbao by Gehry, New York by Wright)---by issuing a \emph{second} \texttt{visual\_search} on the background, so the P--K alternation is functionally necessary rather than a scripted pattern; (3)~each gold entity ($e_1$--$e_4$) is grounded at a distinct step, letting HaPO assign step-level credit at four separate anchors. A trajectory-level reward would collapse all four hops into a single success/failure signal, whereas HaPO rewards the intermediate disambiguation ($e_3$) independently---precisely the step where a standard GRPO agent tends to skip the background verification and answer for the wrong branch.

\end{document}